\pdfoutput=1
\documentclass[]{fairmeta}

\usepackage{amssymb}
\usepackage{amsmath}
\usepackage{colortbl}
\usepackage{tabularx}
\usepackage{float}
\usepackage{wrapfig}
\usepackage{tikz}
\usetikzlibrary{
  arrows.meta,
  positioning,
  fit,
  backgrounds,
  decorations.pathreplacing,
  calc,
  shapes.geometric
}

\tikzset{
  snowflake/.pic={
    \foreach \a in {90,150,...,390} {
      \draw[line width=0.45pt, line cap=round] (0,0) -- (\a:1.5mm);
      \draw[line width=0.45pt, line cap=round] (\a:0.9mm) -- +(\a+45:0.5mm);
      \draw[line width=0.45pt, line cap=round] (\a:0.9mm) -- +(\a-45:0.5mm);
    }
  },
  flame/.pic={
    \fill[x=1mm, y=1mm] (0,-1.5)
      .. controls (1.35,-1.5) and (1.25,0.1) .. (0.15,1.6)
      .. controls (0.25,0.7) and (-0.25,0.5) .. (-0.35,0.2)
      .. controls (-0.55,0.45) and (-0.6,0.7) .. (-0.55,0.9)
      .. controls (-1.3,0.3) and (-1.3,-1.5) .. (0,-1.5) -- cycle;
  },
}
\DeclareRobustCommand{\snowflakeicon}{\tikz[baseline=-0.6ex, scale=0.8]\pic[seqblue]{snowflake};}
\DeclareRobustCommand{\flameicon}{\tikz[baseline=-0.6ex, scale=0.8]\pic[seqorange]{flame};}

\tcbuselibrary{skins,listings}
\newtcblisting{promptbox}[1]{
  listing only,
  title={#1},
  fonttitle=\bfseries\footnotesize,
  coltitle=black,
  colbacktitle=black!8,
  colback=black!3,
  colframe=black!35,
  boxrule=0.6pt,
  arc=3pt,
  left=4pt,
  right=4pt,
  top=2pt,
  bottom=2pt,
  listing options={
    basicstyle=\scriptsize\ttfamily,
    breaklines=true,
    breakindent=0pt,
    columns=fullflexible,
    keepspaces=true,
    showstringspaces=false,
    aboveskip=0pt,
    belowskip=0pt,
  },
}

\definecolor{seqblue}{HTML}{2F6FC4}
\definecolor{seqorange}{HTML}{D9822B}
\definecolor{seqgreen}{HTML}{2E7D32}
\definecolor{seqgray}{HTML}{9AA0A6}
\definecolor{ctxgreen}{HTML}{48B472}
\definecolor{indgray}{HTML}{8C8C8C}
\definecolor{opuscolor}{HTML}{CC785C}
\definecolor{gptcolor}{HTML}{6E6E80}
\definecolor{deepseekcolor}{HTML}{4D6BFE}

\newcommand{\cmark}{{\color{green!55!black}\checkmark}}
\newcommand{\xmark}{{\color{red!70!black}\ensuremath{\times}}}
\newcommand{\opus}{Opus 4.6}
\newcommand{\gpt}{GPT-5.4}
\newcommand{\dps}{DeepSeek v4 Pro}
\newcommand{\muse}{Muse Glimmer}
\newcommand{\gaia}{GAIA2}
\newcommand{\swe}{SWE-Bench Pro}

\AtBeginDocument{%
}

\hypersetup{
  pdfencoding=auto,
  pdftitle={No Reset: Evaluating and Training Persistent Agents on Task Sequences},
  pdfauthor={Romain Froger, Matteo Bettini, Th\'eophile Laurent, Kavya Kopparapu, Gr\'egoire Mialon, Amine Benhalloum, Djam\'e Seddah, Thomas Scialom},
}

\title{No Reset: Evaluating and Training Persistent Agents on Task Sequences}

\author[1,2,3,*]{Romain Froger}
\author[1,*]{Matteo Bettini}
\author[1,*]{Théophile Laurent}
\author[1,\dagger]{Kavya Kopparapu}
\author[1,\dagger]{Grégoire Mialon}
\author[1,\dagger]{Amine Benhalloum}
\author[2,\dagger]{Djamé Seddah}
\author[1,\dagger]{Thomas Scialom}

\affiliation[1]{Meta Superintelligence Labs}
\affiliation[2]{Inria Paris}
\affiliation[3]{Sorbonne Université}

\contribution[*]{Equal contribution}
\contribution[\dagger]{Equal direction}

\abstract{
AI agents are typically evaluated one task at a time, yet deployed agents work through task after task in persistent sessions where conversation history, memory, and files carry over. We build sequential versions of \gaia{} and \swe{} in which agents solve up to 40 tasks in a single session, with tasks and graders unchanged. Task success then depends on the model: \opus{} and \gpt{} stay near their independent baselines, whereas \dps{} loses 5.9 percentage points on average. Behavior changes for every model, with up to 60\% fewer output tokens and less exploration. Agents reuse scripts and knowledge from earlier tasks, but they also carry unchecked assumptions into later ones. We introduce Sequential RL (Seq-RL), which trains agents inside persistent sessions, where the same tasks yield a much stronger training signal than in isolation. Trained only on \gaia{}-style data, Seq-RL narrows the sequential gaps on both \gaia{} splits and outperforms single-task RL trained on the same data. On \swe{}, a domain and harness it never saw in training, it reduces the gap to 6.0 percentage points, against 21.9 for single-task RL. \textbf{Managing accumulated experience is a distinct capability from solving isolated tasks, and training without resets directly improves it.}
}

\date{November 30, 2026}
\correspondence{\email{rfroger@meta.com}}
\metadata[Project page]{\url{https://romainfrog.github.io/no-reset/}}

\begin{document}

\maketitle

\begin{figure}[H]
  \centering
  \includegraphics[width=0.95\linewidth]{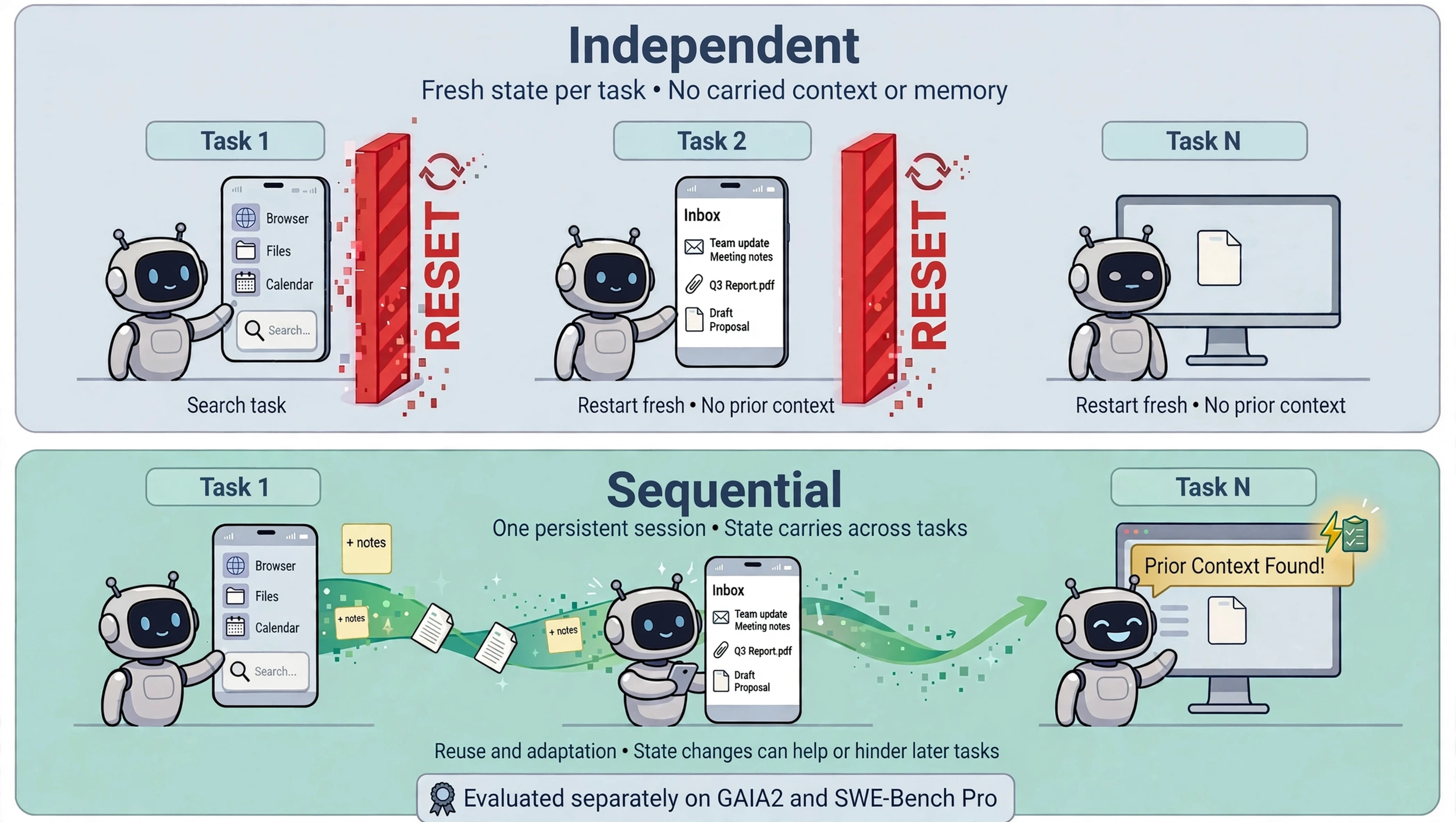}
  \caption{
  \textbf{Independent versus sequential agent evaluation.} The same tasks run in fresh or persistent sessions. Conversation and files carry over across tasks. \gaia{} also preserves memory and environment state, whereas \swe{} checks out each task's starting commit.}
  \label{fig:delta_overview}
\end{figure}

\section{Introduction}
\label{section:intro}

Today's agents operate in persistent sessions: coding assistants such as
Claude Code and Codex, and personal agents such as OpenClaw
\citep{openclaw2026} and Hermes Agent \citep{hermesagent2026}, solve task
after task while conversation history, files, and memory accumulate. What an
agent leaves behind shapes its next task. In one \gaia{} sequence, an agent
writes a script that fetches contacts ten at a time and stops at 130 records.
Because the script stays in the home directory, later tasks reuse it together
with its hidden assumption. Across three frontier models, fixed-bound
pagination scripts succeed 9.5 percentage points less often than scripts that
check they retrieved every record.

Evaluation and training have not followed this shift. Standard agent benchmarks
reset the session between tasks
\citep{jimenez2024swebenchlanguagemodelsresolve,
froger2026gaia2benchmarkingllmagents,deng2025swebenchproaiagents}, and agent
RL typically samples each task from a fresh context. Long-context and
multi-turn studies show that accumulated history degrades performance
\citep{liu2023lostmiddlelanguagemodels,du2025contextlengthhurtsllm} and
reliability \citep{laban2026llms}, but within a single task. Memory and
continual-learning benchmarks compose tasks explicitly designed for cross-task
learning \citep{he2026memoryarenabenchmarkingagentmemory,
dou2025evalearnquantifyinglearningcapability,
asawa2026continuallearningbenchevaluating}. A reset discards the exact state whose reuse or misuse decides deployment
behavior. Whether agents can handle sequences of ordinary tasks, and whether
training can improve that ability, therefore remains open.

We remove the reset between tasks and call this setting sequential evaluation
(\autoref{fig:delta_overview}). In sequential \gaia{}
\citep{froger2026gaia2benchmarkingllmagents}, an agent solves up to 20 tasks
in one session while conversation history, memory, files, and environment state
persist. In sequential \swe{} \citep{deng2025swebenchproaiagents}, it resolves
up to 40 repository issues in commit order: its conversation and home
directory persist, while the repository is checked out at each issue's starting commit.
Tasks and graders stay unchanged, and sequences remain jointly feasible under
intended solutions, so every task compares directly with its independent run.
Unlike ChainWorld \citep{siu2026chainworldcomposinglonghorizondesktop}, which
composes two to four OSWorld tasks, sequential evaluation reuses existing
benchmarks directly over sequences an order of magnitude longer.

Persistence changes how every model works, even when its accuracy barely moves.
\opus{} and \gpt{} stay near their independent baselines, whereas \dps{} loses
5.9 percentage points on average and degrades in every setting. All three
models use up to 60\% fewer output tokens than in independent runs. On \swe{}
they inspect fewer files and edit sooner, while on \gaia{} they write Python
scripts (up to $4\times$ higher share of tool calls) that stay on disk for
later tasks. Reuse saves effort, yet carries unchecked assumptions
forward. Besides the contacts script, \dps{} commits early to broad edits on
\swe{} and modifies existing test files ten times more often in sequence than
in isolation.

Persistent sessions also change what an agent learns from the same data. When
\muse{}-30B is trained with RL on disjoint \gaia{}-style tasks in isolation,
its mean training reward rises by 3\%. When those same tasks follow a long
prior session history, the reward rises by 91\%. Sequential RL (Seq-RL) targets this
setting by continuing a frozen 100K-token session with six tasks and rewarding
each task for its own outcome and downstream outcomes. Unlike methods that
train within a single task
\citep{li-etal-2026-mitigating,li2026compactionrlreinforcementlearningcontext}
or rely on explicit skill libraries and curricula
\citep{wang2026reinforcementlearningselfimprovingagent,
yao2026skillriseagenticreinforcementlearning}, Seq-RL trains the base agent
directly across independent tasks in one session.

Trained only on disjoint \gaia{}-style data, Seq-RL outperforms single-task RL
trained on the same data under sequential evaluation on all three test sets
(\autoref{tab:seq-rl-avgk}) and narrows the sequential gaps on both \gaia{}
splits. Zero-shot on \swe{}, a domain and harness unseen in training, it cuts
the gap to 6.0 percentage points against 21.9 for single-task RL. Seq-RL
preserves the efficiency of persistence while restoring verification: on \swe{}
it still uses about half its independent output tokens, yet averages 2.7 test
turns per task against about 1.1 for the base model and single-task RL. Its
gains also extend to MemoryArena
\citep{he2026memoryarenabenchmarkingagentmemory}, where progress improves in all
four domains. Evaluating and training agents without resets targets a distinct
capability that episodic benchmarks miss, and our protocol applies to
any benchmark whose tasks share an environment.

\section{Related work}
\label{sec:related}
\paragraph{Long-context and multi-turn degradation.}
Long-context evaluations show that performance declines with prompt length
\citep{liu2023lostmiddlelanguagemodels,du2025contextlengthhurtsllm}, and
multi-turn interactions lower reliability as requirements accumulate
\citep{laban2026llms}. Agentic studies report similar failures over long
histories \citep{chung2025evaluatinglongcontextreasoningllmbased,
fang2026agentlongbenchcontrollablelongbenchmark,
sobal2026staminabenchstresstestingcodingagents}, while SelfCompact adapts when
to compact via a training-free rubric
\citep{li2026selfcompactinglanguagemodelagents}. All examine context within a
single task or evolving objective. Here, we evaluate sequences of independent
tasks in one session, where agents must decide what state to reuse, verify, or
ignore.

\begin{table}[!b]
  \centering
  \caption{\textbf{No Reset combines persistent task sequences on fixed standard benchmarks with a training method.} \protect\cmark{} / \protect\xmark{} indicate persistent sessions, agentic environments, unchanged standard tasks and graders (enabling paired independent-versus-sequential comparison), and a training method.}
  \label{tab:related-work}
  \footnotesize
  \setlength{\tabcolsep}{5.5pt}
  \renewcommand{\arraystretch}{1.15}
  \begin{tabularx}{\linewidth}{@{}Xlcccc@{}}
    \toprule
    \textbf{Benchmark} & \textbf{Setting} & \textbf{Persistent} & \textbf{Agentic} & \textbf{Fixed std.\ tasks} & \textbf{Training} \\
    \midrule
    Episodic {\scriptsize(GAIA, $\tau$-bench, SWE-bench)} & Tools \& code & \xmark & \cmark & \cmark & \xmark \\
    Dependent SWE {\scriptsize(Milestone, Chain, EVO, CI)} & Evolving code & \cmark & \cmark & \xmark & \xmark \\
    EvaLearn \citep{dou2025evalearnquantifyinglearningcapability} & Text problems & \cmark & \xmark & \xmark & \xmark \\
    CL-Bench \citep{asawa2026continuallearningbenchevaluating} & 6 stateful domains & \cmark & \cmark & \xmark & \xmark \\
    MemoryArena \citep{he2026memoryarenabenchmarkingagentmemory} & 4 agent domains & \cmark & \cmark & \xmark & \xmark \\
    ChainWorld \citep{siu2026chainworldcomposinglonghorizondesktop} & Desktop GUI {\scriptsize(2--4 tasks)} & \cmark & \cmark & \cmark & \xmark \\
    \rowcolor{seqblue!12}
    \textbf{No Reset} & \textbf{Assistant \& code} {\scriptsize\textbf{(up to 40 tasks)}} & \cmark & \cmark & \cmark & \cmark \\
    \bottomrule
  \end{tabularx}
\end{table}

\paragraph{From episodic benchmarks to sequential task evaluation.}
Most agent benchmarks reset between tasks
\citep{yao2024taubenchbenchmarktoolagentuserinteraction,
mialon2023gaiabenchmarkgeneralai,
jimenez2024swebenchlanguagemodelsresolve,sun2026agentsexam}. Recent coding
benchmarks instead sequence dependent tasks over an evolving codebase to study
error propagation \citep{deng2026swemilestoneevaluatingaiagents,
lam2026swechainbenchmarkingcodingagents,
thai2026sweevobenchmarkingcodingagents,
shastry2026isolatedtasksframeworkevaluating,
chen2026swecievaluatingagentcapabilities}, which couples outcomes to prior
patches and prevents a paired comparison with independent runs. Closest to our
setup, ChainWorld \citep{siu2026chainworldcomposinglonghorizondesktop}
composes two to four atomic OSWorld tasks into compatible sequences with unchanged
evaluators and compares single-turn with multi-turn delivery
(\autoref{tab:related-work}). Our evaluation adapts \gaia{}
\citep{froger2026gaia2benchmarkingllmagents} and \swe{}
\citep{deng2025swebenchproaiagents} into persistent sessions an order of
magnitude longer, with up to 20 and 40 tasks per session. These runs reach
histories up to 900K tokens with repeated compaction, while fixed tasks and
graders allow a direct independent-versus-sequential comparison.

\paragraph{Memory and continual-learning benchmarks.}
A parallel line of benchmarks builds sequences specifically to test whether
agents retain and exploit prior experience
\citep{zheng2025lifelongagentbenchevaluatingllmagents,
he2026memoryarenabenchmarkingagentmemory,
shu2026agentclrigorousevaluationcontinual,
jiang2026seaevalbenchmarkevaluatingselfevolving}. CL-Bench designs tasks around
a learnable latent structure that a stateful system can discover online but a
stateless one cannot \citep{asawa2026continuallearningbenchevaluating}.
EvaLearn groups related text problems by task type with optional rubrics or
teacher feedback \citep{dou2025evalearnquantifyinglearningcapability}, while
ContinualSkillBench links subtasks with explicit skill maintenance
\citep{guan2026continualskillbenchllmagentstruly} and sequential embodied QA
targets memory architecture across questions
\citep{cai2026episodicevaluationmemoryarchitectural}. Because these suites
construct sequences specifically to reward cross-task learning or memory
recall, they cannot tell whether ordinary session persistence helps or hurts on
standard tasks solvable in isolation. Sequential evaluation isolates that effect and
complements memory benchmarks like MemoryArena, which we use as an
out-of-distribution test for Seq-RL (\autoref{sec:seqrl-generalization}).

\paragraph{Training for long-horizon and cross-task operation.}
Recent training methods target cross-task reuse via generated tools, workflow
memories, or skill libraries \citep{cai2024largelanguagemodelstool,
wang2025agentworkflowmemory,wang2026reinforcementlearningselfimprovingagent},
progressive task specification \citep{chen2026conversationllmsteachclose},
explicit context updates \citep{chen2026connectdotstrainingllms}, or curricula
\citep{yao2026skillriseagenticreinforcementlearning}. Closer to our RL setup,
CompactionRL jointly optimizes task execution and within-task compaction
\citep{li2026compactionrlreinforcementlearningcontext}, Learning to Remember
trains memory updates from downstream queries
\citep{zhang2026learningrememberendtoendtraining}, and RLAAR applies single-task
multi-turn RL with abstention rewards \citep{li-etal-2026-mitigating}. Seq-RL
instead adds no memory module or skill library: it trains the
base agent on persistent sequences of independent tasks with local and
downstream rewards, and its gains transfer across domains and harnesses.

\section{Experimental setup}
\label{sec:setup}
\label{sec:seq-eval-setup}
In \textbf{independent evaluation}, each task starts in a fresh session with no
prior conversation or home filesystem state. In \textbf{sequential
evaluation}, each sequence starts clean, but its tasks share one persistent
session whose agent state is never reset. This comparison measures the
end-to-end effect of session persistence under each benchmark's state
semantics, while keeping task definitions and grading unchanged (see
\autoref{fig:delta_overview}, with more implementation details in Appendix
\autoref{fig:eval-setup}).

\paragraph{Benchmarks.}
We reuse independently defined and graded tasks grouped by a shared
environment: a user universe in \gaia{} or a Git repository in \swe{}.
\begin{itemize}
  \item \textbf{\gaia{}} \citep{froger2026gaia2benchmarkingllmagents} evaluates
  personal assistants in a simulated smartphone environment containing email,
  messaging, calendar, and contact applications. We evaluate the \textit{Search}
  (160 tasks) and \textit{Execution} (95 of 160 causally compatible tasks)
  splits, organized into 20 sequences of 4--20 tasks across \emph{universes}
  with shared application state. Tasks are ordered so that completing an earlier
  task as intended never invalidates a later one, whereas agent errors persist.
  Each sequence runs in one OpenClaw~\citep{openclaw2026} session, with
  conversation, memory, home files, and application state all persisting.
  \item \textbf{\swe{}} \citep{deng2025swebenchproaiagents} contains GitHub
  issues from 11 repositories. We group issues by repository, order them by Git
  history, and retain up to 40 per repository for 420 tasks in 11 sequences.
  Each sequence runs in one Mini-SWE-Agent~\citep{yang2024sweagent} session,
  minimally extended with persistent sessions, home-directory persistence, and
  full-history summarization (\autoref{app:mini-swe}). Before each task, the repository
  is checked out to that task's starting commit (with scrubbed Git history and
  no internet access) and the agent is notified, while conversation and
  agent-home state persist.
\end{itemize}
Further task-bundling details, sequence-length distributions, and harness
modifications appear in \autoref{app:adapting}.

\paragraph{Evaluation protocol.}\label{sec:protocol}
We evaluate \gpt{}, \opus{}, and \dps{} under both evaluation modes, using $K=5$
rollouts per \gaia{} task and $K=3$ per \swe{} task. Each sequential rollout is
initialized independently and executes a complete sequence end to end. Both
harnesses use a 200K-token context window by default (summarizing older history near the
limit, while OpenClaw also compacts old tool outputs, \autoref{app:mini-swe}) and all
models use high reasoning effort. Infrastructure failures (e.g., VM crashes or
model API outages) are replayed so the failed attempt does not affect later
tasks.

\paragraph{Metrics.}
Our primary accuracy metric is $\mathrm{avg@}K$, the mean success rate across
$K$ rollouts, reported with 95\% sequence-level cluster-bootstrap confidence
intervals (\autoref{app:eval-stats}). We report $\mathrm{pass@}K$ (coverage)
and $\mathrm{pass}^{K}$ (all-run success) as secondary metrics in
\autoref{app:task-performance}. For each metric $m$, the
\emph{sequential gap} is $\Delta_m = m_{\mathrm{seq}} - m_{\mathrm{ind}}$
(in percentage points, pp), where negative values indicate degradation under
sequential evaluation. Our primary measure of agent effort is output tokens,
alongside LLM turns: a single turn may range from tens to thousands of tokens
depending on the model's reasoning depth and solving strategy, so output tokens
better reflect total effort. On \gaia{}, we also report \emph{Python share}
(fraction of tool calls executing Python) and \emph{Python use} (fraction of
task runs calling Python at least once).

\section{Sequential evaluation results}
\label{sec:results}
We evaluate how session persistence changes task performance
(\autoref{sec:results-accuracy}), how it changes agent workflows
(\autoref{sec:behavior}), and why tuning the context window alone cannot
resolve the resulting trade-offs (\autoref{sec:mechanisms}).

\subsection{Task performance under persistence}
\label{sec:results-accuracy}

\begin{figure}[t]
  \centering
  \includegraphics[width=\textwidth]{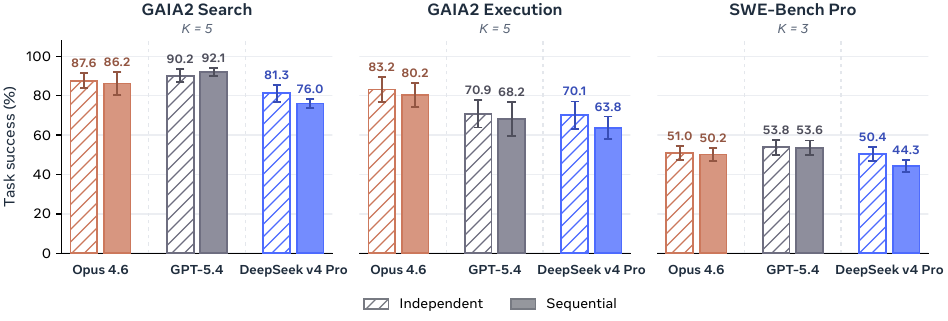}
  \caption{\textbf{Models matched in isolation diverge under persistence: \dps{} drops across every setting while \opus{} and \gpt{} hold steady.}
  Bars show independent (hatched) versus sequential (solid) $\mathrm{avg@}K$ at a 200K-token context window, with 95\% sequence-level cluster-bootstrap confidence intervals.}
  \label{fig:seq-gain}
\end{figure}

Episodic scores hide how differently models handle persistent sessions
(\autoref{fig:seq-gain}). In isolation, \opus{}, \gpt{}, and \dps{} perform
similarly on \swe{} (50.4--53.8\% $\mathrm{avg@}3$), and \gpt{} and \dps{}
nearly tie on \gaia{} Execution (70.9 vs.\ 70.1\%). Removing resets splits them
apart: \opus{} and \gpt{} stay within 1.7 and 0.3~pp of their independent
baselines on average, whereas \dps{} drops by 5.3--6.3~pp across all three
settings ($-5.9$~pp average). Because this loss recurs across two distinct
domains and harnesses, managing persistent state is a model-level capability
rather than a benchmark artifact.

Even when mean accuracy holds steady, reliability drops
(\autoref{tab:abs-gaia2}, \autoref{tab:abs-swe}). Across all three frontier
models, $\mathrm{pass@}K$ barely changes, so persistence rarely makes a
task unsolvable. All-run consistency ($\mathrm{pass}^{K}$), however, falls in
8 of 9 model-setting pairs: even on \swe{}, where \gpt{}'s $\mathrm{avg@}3$
moves by only $-0.2$~pp, its $\mathrm{pass}^{3}$ drops 7.6~pp, and \dps{} loses
8.4--16.3~pp across settings. Persistence hurts run-to-run consistency even
when a model's average score looks unaffected.

\subsection{Behavioral changes under sequential evaluation}
\label{sec:behavior}

\begin{figure}[t]
  \centering
  \includegraphics[width=\textwidth]{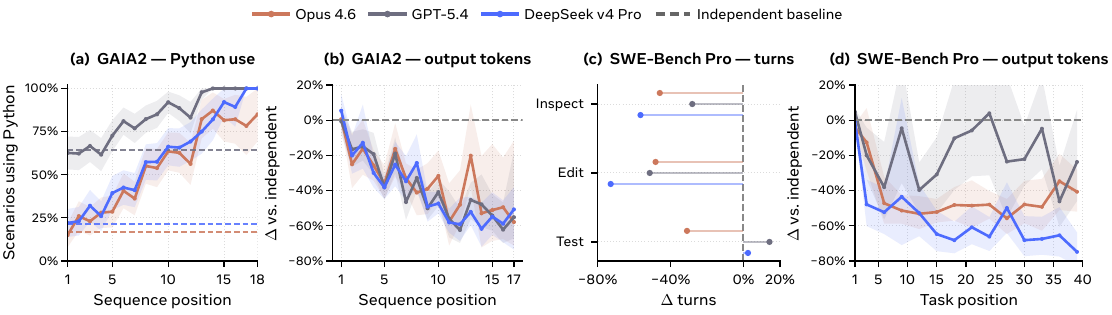}
  \caption{\textbf{Persistence reshapes how every model works, even when its average accuracy barely changes.}
  (a)~\gaia{} Python use by sequence position. (b--d)~Changes relative to independent evaluation in \gaia{} output tokens (b), \swe{} turns by action type (c), and \swe{} output tokens (d).}
  \label{fig:behavioral-adaptation}
\end{figure}

\paragraph{Effort drops before context pressure.}
Every frontier model takes shorter trajectories under persistence: output
tokens drop by 20--60\% and LLM turns by 26--51\% across both benchmarks
(\autoref{fig:behavioral-adaptation}(b,d), \autoref{tab:swe-actions},
\autoref{fig:eff-32k}). On \gaia{}, this reduction begins at the second task
(\autoref{fig:eff-gaia2}), even though full-history summarization never
triggers at 200K (\autoref{tab:compaction}). Because every task is paired with
its own independent run, shorter trajectories reflect an active behavioral
response to carried context rather than easier late tasks or context-window
pressure. Note that these savings measure output work rather than total
inference cost, since sequential prompts carry longer input histories.

\paragraph{\gaia{}: reusable scripts and carried assumptions.}
On \gaia{}, models respond to accumulating history by moving data processing
into persistent Python scripts: Python's share of tool calls rises up to
$4\times$ at 200K and rises further under a tighter 32K window
(\autoref{fig:behavioral-adaptation}(a), \autoref{tab:python-usage}). Without
being prompted, agents write Python scripts in their home directory to filter
raw tool outputs outside the context window and implement missing compound
queries (\autoref{fig:gaia2-scripts}). Because those scripts persist across
tasks, agents reuse them to save effort, yet any shortcut written into them
persists as well. For example, scripts that paginate with a fixed limit (such
as stopping at 130 contacts, \autoref{fig:gaia2-scripts}(a)) succeed 9.5~pp less
often across 1,574 sequential runs than scripts that check for complete
retrieval (77.2\% vs.\ 86.7\%, \autoref{app:gaia2-scripts}). Though
observational, this gap illustrates the risk of reusing scripts without
re-checking their assumptions.

\paragraph{\swe{}: reusing codebase knowledge versus skipping verification.}
On \swe{}, accumulated repository context shifts effort away from exploration:
all three models inspect 24--32\% fewer files, cut inspection turns by
28--56\%, and reach their first edit sooner while dedicating a larger share of
their remaining turns to testing (\autoref{fig:behavioral-adaptation}(c),
\autoref{tab:swe-actions}). Whether this shortcut helps or hurts depends on how
models verify their changes. \opus{} and \gpt{} reuse codebase structure from prior tasks and skip
re-exploring familiar files without losing accuracy (within 0.8 and 0.2~pp of
baseline). \dps{} cuts effort even harder (60\% fewer
output tokens), commits early to broad, poorly localized edits, and rewrites
existing test files $10\times$ more often than in isolation (542 vs.\ 56 edits,
\autoref{app:swe-actions}) to accommodate those changes. Differences in compacted-history training and weaker long-context retrieval
from sparse attention (MRCR 83.5 vs.\ 92.9 for \opus{} at 1M,
\citealp{deepseekai2026deepseekv4highlyefficientmilliontoken}) may contribute.

\subsection{Context length, compaction, and adaptation}
\label{sec:mechanisms}

\begin{figure}[t]
  \centering
  \includegraphics[width=0.82\textwidth]{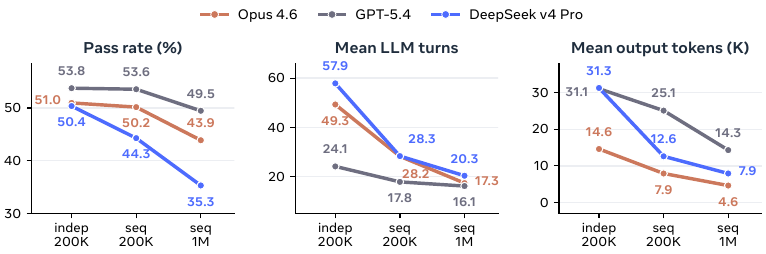}
  \caption{\textbf{Extending the \swe{} context window to 1M tokens worsens sequential accuracy for every frontier model.}
  Independent evaluation at 200K (equivalent to 1M, as independent runs stay well below 200K) versus sequential evaluation at 200K and 1M in $\mathrm{avg@}3$, LLM turns, and output tokens. Position curves and context trajectories appear in \autoref{app:ablations}.}
  \label{fig:swe-1m-main}
\end{figure}

A natural objection is that sequential degradation is a context-window
limitation, fixable by enlarging the window or compacting earlier. Window
ablations rule out both (\autoref{app:ablations}). Expanding \swe{} to a 1M
window so sessions retain up to 900K tokens of raw history worsens sequential
$\mathrm{avg@}3$ for \emph{every} model (\autoref{fig:swe-1m-main}): \gpt{}
($-4.3$~pp) and \opus{} ($-7.1$~pp) drop below baseline, while \dps{} loses
15.1~pp and effort falls further. Information loss from compaction is therefore
not the root cause, since long raw histories hurt performance
\citep{liu2023lostmiddlelanguagemodels,du2025contextlengthhurtsllm}.

Compacting earlier creates the opposite failure. Shrinking \gaia{} to 32K
forces frequent summarization, cuts independent accuracy by 2--12~pp
(\autoref{tab:abs-gaia2-32k}), and makes two models emit \emph{more} output
tokens sequentially as they regenerate discarded context
(\autoref{fig:eff-32k}). In 200K \swe{} runs, summarization also resets part
of the adaptation benefit: output effort rebounds in the two tasks after each
summary (29\% to 46\% of independent tokens for \dps{},
\autoref{app:ablations}) without recovering pass rate. Together, these
ablations show three competing effects: long-context degradation, lossy
compaction, and adaptation to carried history. Tuning the context window only
trades them off, so improving persistent agents requires training inside
persistent sessions (\autoref{sec:seq-rl}).

\section{Seq-RL: training for persistent sessions}
\label{sec:seq-rl}

Standard agent RL trains on isolated tasks from fresh context and never exposes
models to trajectory-dependent histories, though recent work targets
long-lifecycle or cross-task learning
\citep{chen2026connectdotstrainingllms,
yao2026skillriseagenticreinforcementlearning}. We introduce \emph{Seq-RL} and
train \muse{}-30B \citep{abtahizadeh2026museglimmer30bondevice}
directly on task sequences inside persistent sessions.
\subsection{Sequential reinforcement learning setup}
\label{sec:seq-rl-setup}

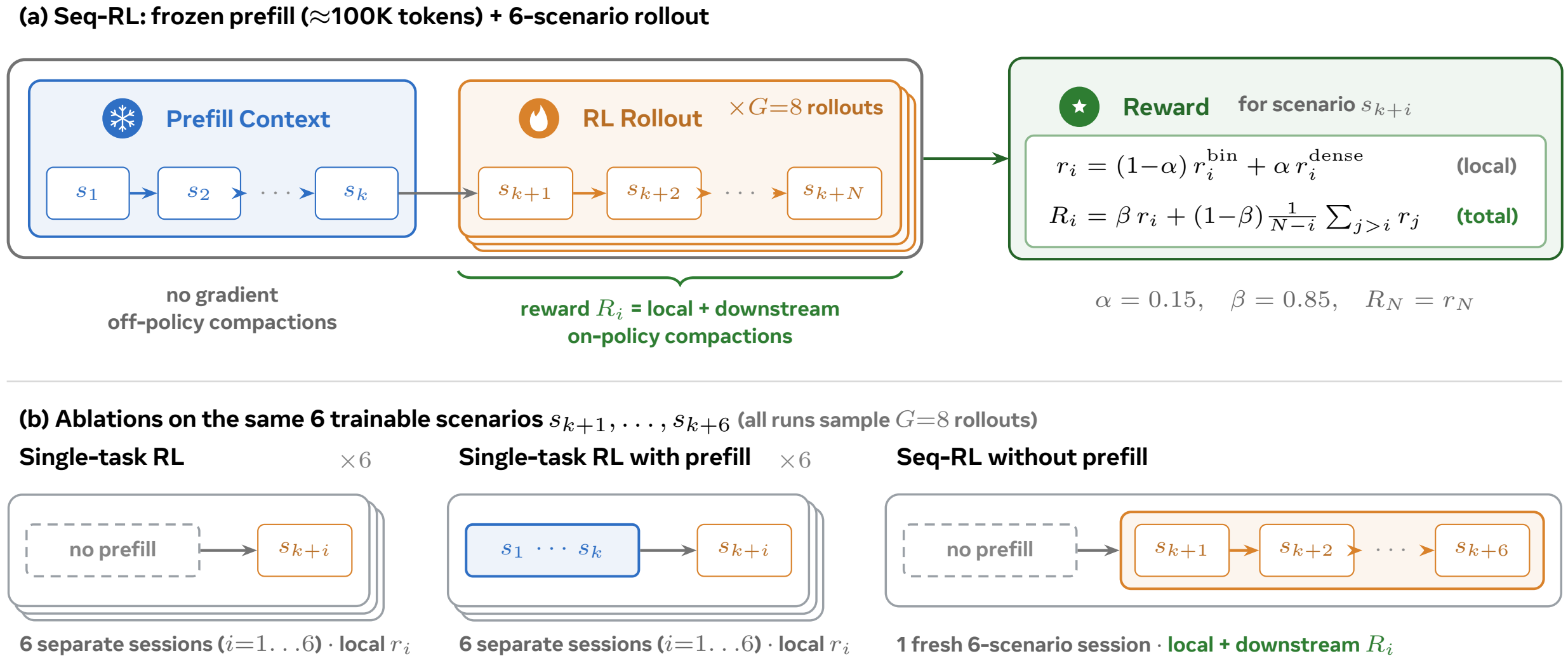
\begin{figure}[!ht]
  \centering
  \begin{tikzpicture}[
    font=\sffamily,
    hdr/.style={font=\sffamily\bfseries\footnotesize},
    ann/.style={font=\sffamily\scriptsize, text=black!60},
    dots/.style={font=\sffamily\scriptsize, text=black!45},
    pill/.style={rounded corners=2pt, draw, minimum height=5.4mm, inner xsep=2pt, inner ysep=1pt, font=\sffamily\scriptsize},
    pinner/.style={pill, minimum width=8.6mm, draw=seqblue, fill=white, text=seqblue},
    rinner/.style={pill, minimum width=9.8mm, draw=seqorange, fill=white, text=seqorange!85!black},
    pfx/.style={pill, minimum width=18mm, draw=seqblue, fill=seqblue!8, text=seqblue, line width=0.75pt},
    fresh/.style={pill, minimum width=18mm, draw=seqgray, densely dashed, fill=white, text=black!48, line width=0.65pt},
    barr/.style={-{Stealth[length=1.6mm]}, line width=0.8pt, seqblue},
    oarr/.style={-{Stealth[length=1.6mm]}, line width=0.8pt, seqorange},
    garr/.style={-{Stealth[length=1.6mm]}, line width=0.8pt, black!55},
    badgeb/.style={circle, fill=seqblue, minimum size=4.2mm, inner sep=0pt},
    badgeo/.style={circle, fill=seqorange, minimum size=4.2mm, inner sep=0pt},
    badgeg/.style={circle, fill=seqgreen, minimum size=4.2mm, inner sep=0pt},
    card/.style={rounded corners=4pt, draw=seqgray, line width=0.65pt, fill=white},
    cont/.style={rounded corners=3pt, draw=seqorange, line width=0.75pt, fill=seqorange!8},
    rewardstar/.pic={\node[star, star points=5, star point ratio=2.15, fill=white, inner sep=0.75pt] {};},
  ]

  \node[pinner] (p1) at (0.82, 0) {$s_1$};
  \node[pinner] (p2) at (1.98, 0) {$s_2$};
  \node[dots]   (pd) at (2.80, 0) {$\cdots$};
  \node[pinner] (pk) at (3.62, 0) {$s_k$};
  \draw[barr] (p1)--(p2); \draw[barr] (p2)--(pd); \draw[barr] (pd)--(pk);

  \node[badgeb] (pb) at (1.18, 0.78) {}; \pic[white,scale=0.9] at (pb.center) {snowflake};
  \node[hdr, text=seqblue, anchor=west] (pt) at ([xshift=1.1mm]pb.east) {Prefill Context};

  \node[rinner] (r1) at (5.38, 0) {$s_{k+1}$};
  \node[rinner] (r2) at (6.72, 0) {$s_{k+2}$};
  \node[dots]   (rd) at (7.64, 0) {$\cdots$};
  \node[rinner] (rk) at (8.60, 0) {$s_{k+N}$};
  \draw[oarr] (r1)--(r2); \draw[oarr] (r2)--(rd); \draw[oarr] (rd)--(rk);

  \node[badgeo] (rb) at (5.52, 0.78) {}; \pic[white,scale=0.9] at (rb.center) {flame};
  \node[hdr, text=seqorange!85!black, anchor=west] (rt) at ([xshift=1.1mm]rb.east) {RL Rollout};

  \begin{scope}[on background layer]
    \node[fit=(p1)(pk)(pb)(pt), inner xsep=5pt, inner ysep=5pt] (lpt) {};
    \node[fit=(r1)(rk)(rb)(rt), inner xsep=5.5pt, inner ysep=5pt] (rfit) {};
    \node[rounded corners=5pt, draw=black!55, line width=0.9pt, fill=white,
          fit=(lpt)(rfit), inner xsep=6pt, inner ysep=6pt] (ctx) {};
    \node[rounded corners=3.5pt, draw=seqblue, line width=0.75pt, fill=seqblue!8,
          fit=(p1)(pk)(pb)(pt), inner xsep=5pt, inner ysep=5pt] (lp) {};
    \foreach \s in {2,1}
      \draw[rounded corners=3.5pt, draw=seqorange, line width=0.7pt, fill=seqorange!4]
        ([xshift=\s*2pt, yshift=-\s*2pt]rfit.south west) rectangle
        ([xshift=\s*2pt, yshift=-\s*2pt]rfit.north east);
    \node[rounded corners=3.5pt, draw=seqorange, line width=0.75pt, fill=seqorange!8,
          fit=(r1)(rk)(rb)(rt), inner xsep=5.5pt, inner ysep=5pt] (rp) {};
  \end{scope}
  \draw[garr] (pk.east) -- (r1.west);
  \node[ann, font=\sffamily\bfseries\scriptsize, text=seqorange!85!black, anchor=north east]
    at ([xshift=-2pt, yshift=-2pt]rp.north east) {$\times G{=}8$ rollouts};

  \node[hdr, anchor=south west] at ([yshift=1.8mm]ctx.north west)
    {(a) Seq-RL: frozen prefill ($\approx$100K tokens) + 6-scenario rollout};

  \coordinate (rwL) at (10.42, 0);
  \coordinate (rwR) at (16.18, 0);
  \draw[rounded corners=4pt, draw=seqgreen!80!black, line width=0.75pt, fill=seqgreen!7]
    (rwL |- ctx.south) rectangle (rwR |- ctx.north);

  \node[badgeg] (gb) at (11.15, 0.90) {}; \pic at (gb.center) {rewardstar};
  \node[hdr, text=seqgreen!85!black, anchor=west] (gt) at ([xshift=1.1mm]gb.east) {Reward};
  \node[ann, anchor=west] (gs) at ([xshift=0.5mm]gt.east) {for scenario $s_{k+i}$};

  \node[
    rounded corners=2.5pt, draw=seqgreen!55, fill=white, line width=0.65pt,
    inner xsep=4.5pt, inner ysep=4pt, font=\sffamily\scriptsize, align=left,
    minimum width=5.42cm
  ] (eqbox) at (13.30, 0.02) {
    $\begin{aligned}
      r_i &= (1{-}\alpha)\,r_i^{\mathrm{bin}} + \alpha\,r_i^{\mathrm{dense}} && \text{\color{black!55}\scriptsize(local)}\\[2.5pt]
      R_i &= \beta\,r_i + (1{-}\beta)\tfrac{1}{N-i}\textstyle\sum_{j>i} r_j && \text{\color{seqgreen}\scriptsize(total)}
    \end{aligned}$
  };
  \node[ann, anchor=north] at ([yshift=-2.0mm]ctx.south -| eqbox)
    {$\alpha=0.15,\quad \beta=0.85,\quad R_N=r_N$};
  \draw[-{Stealth[length=1.8mm]}, line width=0.85pt, seqgreen!80!black]
    (ctx.east) -- (rwL |- ctx.east);

  \node[ann, align=center, anchor=north] at ([yshift=-2.0mm]ctx.south -| lp)
    {no gradient\\off-policy compactions};
  \draw[decorate, decoration={brace, amplitude=4pt, mirror}, seqgreen, line width=0.8pt]
    ([yshift=-3.5pt]ctx.south -| rp.west) -- ([yshift=-3.5pt]ctx.south -| rp.east);
  \node[ann, text=seqgreen, align=center, anchor=north] at ([yshift=-8.5pt]ctx.south -| rp)
    {\textbf{reward $R_i$ = local + downstream}\\on-policy compactions};

  \draw[black!15, line width=0.5pt] (ctx.west |- 0,-1.96) -- (rwR |- 0,-1.96);

  \node[hdr, anchor=north west] at (ctx.west |- 0,-2.12)
    {(b) Ablations on the same 6 trainable scenarios $s_{k+1},\ldots,s_{k+6}$ {\scriptsize\color{black!55}(all runs sample $G{=}8$ rollouts)}};

  \node[fresh]  (f1) at (1.08, -3.72) {no prefill};
  \node[rinner] (a1) at (3.08, -3.72) {$s_{k+i}$};
  \draw[garr] (f1)--(a1);
  \begin{scope}[on background layer]
    \node[fit=(f1)(a1), inner xsep=5pt, inner ysep=8.5pt] (k1t) {};
    \foreach \s in {2,1}
      \draw[card] ([xshift=\s*2pt, yshift=-\s*2pt]k1t.south west) rectangle
                  ([xshift=\s*2pt, yshift=-\s*2pt]k1t.north east);
    \node[card, fit=(f1)(a1), inner xsep=5pt, inner ysep=8.5pt] (k1) {};
  \end{scope}
  \node[hdr, anchor=south west] at ([yshift=1.2mm]k1.north west) {Single-task RL};
  \node[ann, font=\sffamily\bfseries\scriptsize, text=black!50, anchor=south east]
    at ([xshift=4pt, yshift=1.2mm]k1.north east) {$\times 6$};
  \node[ann, anchor=north west] at (k1.west |- 0,-4.48)
    {6 separate sessions ($i{=}1{\ldots}6$) $\cdot$ local $r_i$};

  \node[pfx]    (p1b) at (5.66, -3.72) {$s_1\ \cdots\ s_k$};
  \node[rinner] (a2)  at (7.66, -3.72) {$s_{k+i}$};
  \draw[garr] (p1b)--(a2);
  \begin{scope}[on background layer]
    \node[fit=(p1b)(a2), inner xsep=5pt, inner ysep=8.5pt] (k2t) {};
    \foreach \s in {2,1}
      \draw[card] ([xshift=\s*2pt, yshift=-\s*2pt]k2t.south west) rectangle
                  ([xshift=\s*2pt, yshift=-\s*2pt]k2t.north east);
    \node[card, fit=(p1b)(a2), inner xsep=5pt, inner ysep=8.5pt] (k2) {};
  \end{scope}
  \node[hdr, anchor=south west] at ([yshift=1.2mm]k2.north west) {Single-task RL with prefill};
  \node[ann, font=\sffamily\bfseries\scriptsize, text=black!50, anchor=south east]
    at ([xshift=4pt, yshift=1.2mm]k2.north east) {$\times 6$};
  \node[ann, anchor=north west] at (k2.west |- 0,-4.48)
    {6 separate sessions ($i{=}1{\ldots}6$) $\cdot$ local $r_i$};

  \node[fresh]  (f2) at (10.22, -3.72) {no prefill};
  \node[rinner] (b1) at (12.22, -3.72) {$s_{k+1}$};
  \node[rinner] (b2) at (13.52, -3.72) {$s_{k+2}$};
  \node[dots]   (bd) at (14.42, -3.72) {$\cdots$};
  \node[rinner] (b6) at (15.35, -3.72) {$s_{k+6}$};
  \draw[oarr] (b1)--(b2); \draw[oarr] (b2)--(bd); \draw[oarr] (bd)--(b6);
  \begin{scope}[on background layer]
    \node[fit=(b1)(b6), inner xsep=4pt, inner ysep=3.5pt] (bct) {};
    \node[card, fit=(f2)(bct), inner xsep=5pt, inner ysep=5pt] (k3) {};
    \node[cont, fit=(b1)(b6), inner xsep=4pt, inner ysep=3.5pt] (bc) {};
  \end{scope}
  \draw[garr] (f2)--(bc);
  \node[hdr, anchor=south west] at ([yshift=1.2mm]k3.north west) {Seq-RL without prefill};
  \node[ann, anchor=north west] at (k3.west |- 0,-4.48)
    {1 fresh 6-scenario session $\cdot$ {\color{seqgreen}\textbf{local + downstream $R_i$}}};

  \end{tikzpicture}

  \caption{\textbf{Seq-RL trains on sequential task continuations after a frozen long-context prefill.}
  (a)~\snowflakeicon~The frozen prefill receives no gradient, while \flameicon~the policy samples $G=8$ group rollouts (stacked orange boxes) over the next $N=6$ scenarios under local and downstream rewards ($R_i$).
  (b)~Three ablations over the same six scenarios $s_{k+1},\ldots,s_{k+6}$ isolate the frozen prefill and 6-scenario sequences (all runs sample $G=8$ rollouts). Stacked gray cards denote 6 separate single-scenario sessions ($i=1,\ldots,6$, local reward $r_i$ only), whereas Seq-RL without prefill runs 1 fresh 6-scenario session under the same local and downstream reward $R_i$.}
  \label{fig:seq-rl}
\end{figure}

\paragraph{Training data.}
Following the ARE/\gaia{} annotation protocol \citep{froger2025are}, human
annotators created and verified 27K \gaia{}-style scenarios across universes
disjoint from the evaluation set. We arrange them into 1.6K causally compatible
sequences of up to 50 scenarios (\autoref{app:seqrl-training-details}). Each
sequence yields one training example consisting of a frozen prefill followed by
$N=6$ trainable scenarios, so only 9.6K scenarios receive gradient while the
rest serve as prefill context or go unused.

\paragraph{Seq-RL rollout construction.}
Training on-policy over full 20--50 scenario sequences is prohibitively
expensive, so we expose the policy to long carried contexts at bounded
rollout cost by starting from a frozen prefill. We run the initial
\muse{}-30B checkpoint in a lightweight internal harness with full-history
compaction until it performs at least $c \in \{0,1,2\}$ full-history
compactions and exceeds roughly 100K tokens, then snapshot the complete agent
and environment state at the next scenario boundary $s_k$. Prefills are
generated once and stay fixed during training. Because roughly 25\% of prefill
scenarios succeed, these snapshots expose the policy to realistic failures and
compaction artifacts alongside successes. From each snapshot, the policy
samples $G=8$ independent continuations over the next six scenarios
($s_{k+1},\ldots,s_{k+6}$) without resets at a 128K-token context window
(\autoref{fig:seq-rl}(a)). Only model-generated tokens in these six scenarios,
including on-policy compactions, receive gradient, while prefill tokens and
environment observations are masked.

\paragraph{Reward and optimization.}
Each scenario's local reward $r_i$ mixes binary task success with a dense
$\mathrm{F}_2$ score over write actions ($\alpha=0.15$). Its training reward
$R_i$ combines $r_i$ with the mean local reward of later scenarios in the same
continuation ($\beta=0.85$, $R_N=r_N$). Averaging rather than summing keeps
$R_i$ on a constant scale across positions, and every trainable token in
scenario $i$ (including compaction tokens) receives $R_i$. We optimize with a
critic-free, GRPO-style objective
\citep{shao2024deepseekmathpushinglimitsmathematical} using one baseline per
prefill group (equations and hyperparameters in
\autoref{app:seqrl-training-details}).

\paragraph{Controls.}
To separate long-context exposure from sequential training, we ablate the
frozen prefill and the 6-scenario sequence separately
(\autoref{fig:seq-rl}(b)), plus a $\beta=1$ variant that removes downstream
credit (\autoref{app:seqrl-downstream-ablation}). All runs start from the same
\muse{}-30B checkpoint and train on the exact same 9.6K scenarios (compute and
training details in \autoref{app:seqrl-training-details}). We evaluate all four controls at the final checkpoint for our main accuracy and
effort comparison (\autoref{fig:seqrl-effort}, \autoref{tab:seq-rl-avgk}).
Intermediate-checkpoint, behavioral, and out-of-distribution analyses compare
\muse{}-30B, single-task RL, and Seq-RL.

\subsection{Same data, more signal}
\label{sec:seqrl-training-dynamics}

\begin{figure}[t]
  \centering
  \includegraphics[width=0.92\textwidth]{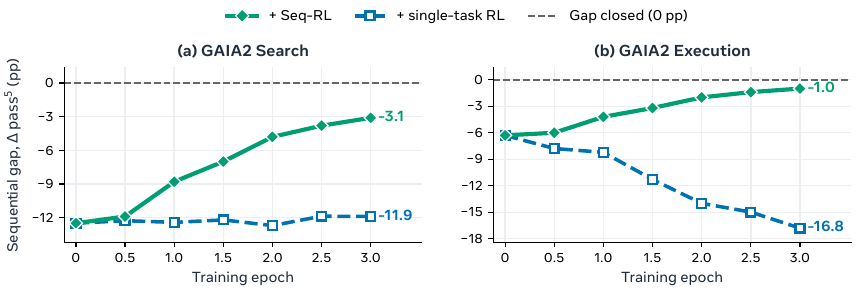}
  \caption{\textbf{Seq-RL closes the held-out \gaia{} consistency gap during training while single-task RL widens it.}
  Sequential gap $\Delta\mathrm{pass}^{5} = \mathrm{pass}^{5}_{\mathrm{seq}} - \mathrm{pass}^{5}_{\mathrm{ind}}$ on \gaia{} Search (a) and Execution (b) across three training epochs.}
  \label{fig:seqrl-gap-growing}
\end{figure}

For a model already post-trained on tool use and agentic environments, the 9.6K
training scenarios leave little room for improvement in isolation: single-task RL starts at
0.537 mean local reward and climbs only 3\% to 0.555 over three epochs
(\autoref{fig:seqrl-reward-curves}). Evaluated inside persistent sessions after
a 100K-token prefill, the exact same scenarios start far harder (0.232) and
climb 91\% to 0.443. The training signal comes from how the data is presented, not from adding new
data.

On held-out \gaia{}, this signal narrows the sequential gap
(\autoref{fig:seqrl-gap-growing}). Tracking $\Delta\mathrm{pass}^{5}$,
which requires all five rollouts to succeed, Seq-RL narrows the gap at every
checkpoint from $-12.5$ to $-3.1$~pp on Search and from $-6.3$ to $-1.0$~pp on
Execution while improving both independent and sequential scores
(\autoref{tab:seq-rl-main}). Single-task RL moves in the opposite direction: it
leaves Search flat ($-11.9$~pp) and widens Execution to $-16.8$~pp because its
independent gains ($\mathrm{pass}^{5}$ 66.3 to 74.7) fail to carry into
persistent sessions (60.0 to 57.9). Downstream credit ($\beta=0.85$ vs.\
$\beta=1$) adds a small further gain (\autoref{app:seqrl-downstream-ablation}).

\subsection{Seq-RL improves agents under persistent context}
\label{sec:seqrl-results}

\begin{figure}[t]
  \centering
  \includegraphics[width=\textwidth]{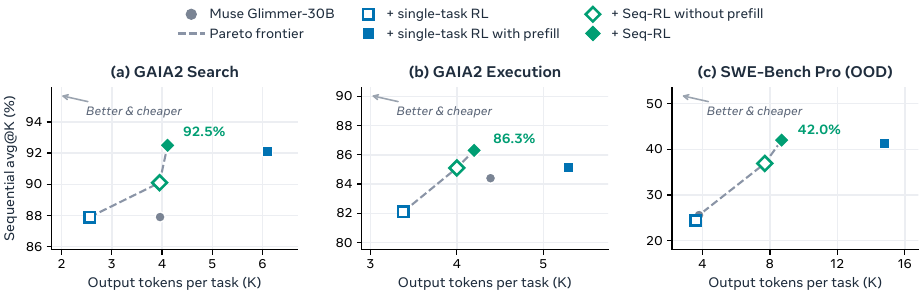}
  \caption{\textbf{Seq-RL achieves the highest sequential accuracy at lower output effort than prefill-only RL.}
  Sequential $\mathrm{avg@}5$ on \gaia{} Search and Execution and $\mathrm{avg@}3$ on \swe{} against mean output tokens per task at a 200K-token window.}
  \label{fig:seqrl-effort}
\end{figure}

\begin{table}[!ht]
  \centering
  \caption{\textbf{Seq-RL achieves the best sequential accuracy and smallest sequential gaps across all three settings.}
  $\mathrm{avg@}5$ on \gaia{} Search and Execution and $\mathrm{avg@}3$ on \swe{} at a 200K-token window (models trained at 128K). \protect\cmark{} / \protect\xmark{} indicate whether training uses the frozen prefill and 6-scenario sequences. $\Delta=\mathrm{Seq.}-\mathrm{Indep.}$, and $\pm$ gives the 95\% cluster-bootstrap half-width in gray (\autoref{app:eval-stats}).}
  \label{tab:seq-rl-avgk}
  \scriptsize
  \setlength{\tabcolsep}{2.5pt}
  \renewcommand{\arraystretch}{1.15}
  \newcommand{\bestscore}[1]{\boldsymbol{#1}}
  \newcommand{\scoreci}[2]{#1{\color{black!50}\scriptstyle\,\pm#2}}
  \begin{tabularx}{\linewidth}{@{}Xcc>{\hspace{5pt}}rrr>{\hspace{5pt}}rrr>{\hspace{5pt}}rrr@{}}
    \toprule
    & \multicolumn{2}{c}{\textbf{Training}}
    & \multicolumn{3}{c}{\textbf{\gaia{} Search}}
    & \multicolumn{3}{c}{\textbf{\gaia{} Execution}}
    & \multicolumn{3}{c}{\textbf{\swe{} (OOD)}} \\
    \cmidrule(lr){2-3}\cmidrule(l{6pt}r{2pt}){4-6}\cmidrule(l{6pt}r{2pt}){7-9}\cmidrule(l{6pt}){10-12}
    \textbf{Method} & \textbf{Prefill} & \textbf{Sequence}
    & \textbf{Indep.} & \textbf{Seq.} & $\Delta$
    & \textbf{Indep.} & \textbf{Seq.} & $\Delta$
    & \textbf{Indep.} & \textbf{Seq.} & $\Delta$ \\
    \midrule
    \muse-30B & -- & --
    & $\scoreci{91.4}{3.3}$ & $\scoreci{87.9}{4.2}$ & $-3.5$
    & $\scoreci{86.1}{4.8}$ & $\scoreci{84.4}{4.3}$ & $-1.7$
    & $\scoreci{45.8}{4.4}$ & $\scoreci{25.6}{4.0}$ & $-20.2$ \\
    \midrule
    + single-task RL & \xmark & \xmark
    & $\scoreci{92.1}{3.3}$ & $\scoreci{87.9}{3.2}$ & $-4.3$
    & $\scoreci{\bestscore{88.6}}{4.6}$ & $\scoreci{82.1}{4.6}$ & $-6.5$
    & $\scoreci{46.3}{4.4}$ & $\scoreci{24.4}{2.9}$ & $-21.9$ \\
    + single-task RL with prefill & \cmark & \xmark
    & $\scoreci{\bestscore{92.8}}{3.1}$ & $\scoreci{92.1}{3.7}$ & $-0.7$
    & $\scoreci{86.3}{3.2}$ & $\scoreci{85.1}{4.2}$ & $-1.2$
    & $\scoreci{47.6}{4.5}$ & $\scoreci{41.3}{4.2}$ & $-6.3$ \\
    + Seq-RL without prefill & \xmark & \cmark
    & $\scoreci{\bestscore{92.8}}{3.1}$ & $\scoreci{90.1}{3.1}$ & $-2.7$
    & $\scoreci{88.2}{4.1}$ & $\scoreci{85.1}{4.2}$ & $-3.1$
    & $\scoreci{47.6}{4.5}$ & $\scoreci{36.9}{3.1}$ & $-10.7$ \\
    \rowcolor{seqgreen!10}
    \textbf{+ Seq-RL} & \cmark & \cmark
    & $\scoreci{\bestscore{92.8}}{3.1}$ & $\scoreci{\bestscore{92.5}}{3.0}$ & $\bestscore{-0.3}$
    & $\scoreci{87.2}{5.2}$ & $\scoreci{\bestscore{86.3}}{6.1}$ & $\bestscore{-0.9}$
    & $\scoreci{\bestscore{48.0}}{4.5}$ & $\scoreci{\bestscore{42.0}}{3.4}$ & $\bestscore{-6.0}$ \\
    \bottomrule
  \end{tabularx}
\end{table}

\paragraph{Sequential gains and zero-shot transfer.}
Seq-RL achieves the highest sequential accuracy and smallest sequential gaps
across all three settings (\autoref{tab:seq-rl-avgk}), and generalizes from
128K training to the 200K evaluation window. Training on isolated tasks does not
help persistent sessions: single-task RL stays at or below \muse{}-30B
sequentially on every setting and finishes last on \gaia{} Execution (82.1)
despite improving independent accuracy (86.1 to 88.6). Our evaluations already
test cross-harness transfer from the lightweight training harness to OpenClaw
on \gaia{}, and \swe{} further tests zero-shot transfer across both domain and
harness (no software-engineering data in RL, evaluated in Mini-SWE-Agent). On
\swe{}, \muse{}-30B suffers a $-20.2$~pp sequential drop, far steeper than the
three frontier models at 200K in \autoref{sec:results} (at worst $-6.3$~pp), and
single-task RL leaves both independent and sequential scores statistically
unchanged. Seq-RL raises sequential $\mathrm{avg@}3$ by 16.4~pp while
independent accuracy barely moves. It cuts the gap to $-6.0$~pp and improves
$\mathrm{pass@}3$ and $\mathrm{pass}^{3}$ (\autoref{tab:seq-rl-main}).
The gain holds in all 11 repositories and still averages +13.1~pp over
\muse{}-30B when excluding the top three (\autoref{tab:seq-rl-swe-repos}).

\paragraph{Prefill recovers accuracy, and sequential training adds efficiency.}
Our controls separate the effect of the frozen prefill from sequential
training (\autoref{fig:seqrl-effort}, \autoref{tab:seq-rl-avgk}).
Adding the prefill alone (single-task RL with prefill) closes almost the entire
accuracy gap, within 0.4--1.2~pp of Seq-RL across all three settings.
Long-context exposure and on-policy compaction drive most of this accuracy
recovery, at the cost of much longer outputs (6.09K output tokens on Search and
14.8K on \swe{}). Training on 6-scenario sequences without a prefill stays concise and
beats single-task RL on \swe{} (36.9 vs.\ 24.4), yet trails Seq-RL in accuracy
on every setting. Combining both components places Seq-RL at the top-left of
\autoref{fig:seqrl-effort}: it achieves the highest sequential $\mathrm{avg@}K$
on all three settings while using 21--41\% fewer output tokens and 18--36\%
fewer turns than prefill-only RL.

\begin{figure}[t]
  \centering
  \includegraphics[width=\textwidth]{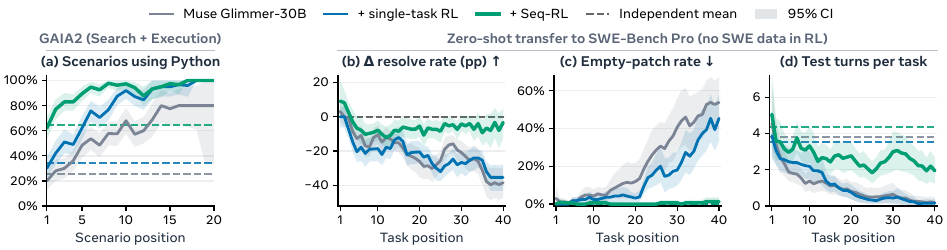}
  \caption{\textbf{Seq-RL sustains verification across long sequences on \gaia{} and \swe{}.}
  (a)~Fraction of \gaia{} scenarios using Python by position (Search and Execution pooled).
  (b--d)~Zero-shot \swe{} resolve rate, empty-patch rate, and test turns (five-position rolling means with 95\% cluster-bootstrap CIs, dashed lines show independent means).}
  \label{fig:seqrl-behavior-position}
\end{figure}

\paragraph{Seq-RL preserves reuse while restoring verification.}
Behavior changes explain how Seq-RL closes the gap
(\autoref{fig:seqrl-behavior-position}). On \gaia{}, Seq-RL uses Python in
nearly every sequential Search run (96.1\% vs.\ 67.0\% for \muse{}-30B, and
80.3\% vs.\ 33.6\% independently, \autoref{tab:seqrl-gaia2-behavior}).
In traces, it uses scripts to recompute fresh task evidence instead of trusting
stale context.
On \swe{}, \muse{}-30B and single-task RL degrade with sequence position:
resolve rate falls, empty patches rise to account for roughly 30\% of the drop,
and test turns approach zero (\autoref{fig:seqrl-behavior-position}(b--d)).
Seq-RL avoids all three trends: it keeps roughly 54\% output-token savings
relative to independent runs while sustaining 2.7 test turns per task
sequentially versus $\approx 1.1$ for the other two checkpoints (and also tests
more independently, 4.4 vs.\ $\approx 3.5$, \autoref{tab:seq-rl-swe-behavior}).

\subsection{Transfer beyond sequential evaluation}
\label{sec:seqrl-generalization}

We next test whether persistent-session RL transfers zero-shot to two unrelated
benchmarks using their official evaluation code (\autoref{tab:seq-rl-ood}):
MRCR v2 \citep{vodrahalli2024michelangelolongcontextevaluations}, which
measures passive long-context needle retrieval, and MemoryArena
\citep{he2026memoryarenabenchmarkingagentmemory}, which tests active memory use
across interdependent sessions in four unseen domains. On MRCR v2, both RL runs improve over \muse{}-30B with overlapping confidence
intervals, and Seq-RL scores highest at 32K, 64K, and 128K. On MemoryArena,
single-task RL changes little and drops 10~pp in formal-physics success (35.0
to 25.0), whereas Seq-RL improves progress in all four domains: formal-physics
success rises 20~pp (to 55.0) and travel-planning progress more than triples
(3.1 to 10.3). Because \autoref{tab:seq-rl-ood} compares against single-task
RL rather than prefill-only RL, these gains reflect the combined effect of
Seq-RL over single-task RL.

\begin{table}[!ht]
  \centering
  \caption{\textbf{Seq-RL transfers zero-shot to MemoryArena across all four domains and improves over \muse{}-30B on MRCR v2.}
  Entries report the mean with 95\% CIs in gray. MRCR v2 uses $\mathrm{avg@}4$ for eight-needle retrieval, and MemoryArena uses $\mathrm{avg@}5$ over five runs for Progress Score (PS) and Success Rate (SR).}
  \label{tab:seq-rl-ood}
  \scriptsize
  \setlength{\tabcolsep}{1.0pt}
  \renewcommand{\arraystretch}{1.08}
  \newcommand{\oodbest}[1]{\boldsymbol{#1}}
  \newcommand{\avgci}[2]{\ensuremath{#1{\color{black!50}\scriptscriptstyle\,\pm\,#2}}}
  \resizebox{\linewidth}{!}{%
  \begin{tabular}{@{}l>{\hspace{6pt}}c*{10}{c}@{}}
    \toprule
      & \multicolumn{3}{c}{\textbf{MRCR v2} (8 needles)}
      & \multicolumn{8}{c}{\textbf{MemoryArena} (concatenated history)} \\
    \cmidrule(l{7pt}r{2pt}){2-4}\cmidrule(lr){5-12}
    \multirow{2}{*}{\textbf{Method}}
      & \multicolumn{3}{c}{avg@4$\uparrow$}
      & \multicolumn{2}{c}{\textbf{Formal math}}
      & \multicolumn{2}{c}{\textbf{Formal physics}}
      & \multicolumn{2}{c}{\textbf{Bundled shopping}}
      & \multicolumn{2}{c}{\textbf{Travel planning}} \\
    \cmidrule(l{7pt}r{2pt}){2-4}\cmidrule(lr){5-6}\cmidrule(lr){7-8}
    \cmidrule(lr){9-10}\cmidrule(lr){11-12}
      & \textbf{32K} & \textbf{64K} & \textbf{128K}
      & \textbf{PS} & \textbf{SR}
      & \textbf{PS} & \textbf{SR}
      & \textbf{PS} & \textbf{SR}
      & \textbf{PS} & \textbf{SR} \\
    \midrule
    \muse-30B
      & \avgci{26.3}{3.0} & \avgci{19.0}{2.7} & \avgci{13.0}{2.2}
      & \avgci{34.9}{1.1} & \avgci{2.0}{1.4}
      & \avgci{60.8}{1.0} & \avgci{35.0}{4.4}
      & \avgci{50.1}{0.8} & \avgci{4.0}{0.6}
      & \avgci{3.1}{0.4} & \avgci{0.1}{0.2} \\
    \midrule
    + single-task RL
      & \avgci{34.7}{3.5} & \avgci{23.8}{2.9} & \avgci{19.5}{2.1}
      & \avgci{35.1}{0.8} & \avgci{\oodbest{3.0}}{1.4}
      & \avgci{58.4}{1.0} & \avgci{25.0}{4.4}
      & \avgci{50.0}{0.8} & \avgci{4.0}{0.6}
      & \avgci{4.9}{0.5} & \avgci{0.4}{0.3} \\
    \rowcolor{seqgreen!10}
    \textbf{+ Seq-RL}
      & \avgci{\oodbest{36.2}}{2.4}
      & \avgci{\oodbest{27.0}}{3.7}
      & \avgci{\oodbest{22.3}}{2.9}
      & \avgci{\oodbest{36.9}}{0.8} & \avgci{\oodbest{3.0}}{1.4}
      & \avgci{\oodbest{71.4}}{1.5} & \avgci{\oodbest{55.0}}{4.4}
      & \avgci{\oodbest{52.9}}{0.8} & \avgci{\oodbest{5.3}}{0.6}
      & \avgci{\oodbest{10.3}}{0.5} & \avgci{\oodbest{0.7}}{0.3} \\
    \bottomrule
  \end{tabular}%
  }
\end{table}

\section{Discussion}

\paragraph{Limitations.}
We measure the end-to-end sequential gap and identify three competing effects
behind it (\autoref{sec:mechanisms}), but we do not causally separate individual error
sources such as environment corruption, lossy compaction, stale beliefs, or
reduced effort. Because these mechanisms interact within the same trajectory,
isolating them requires dedicated interventions. Prior work separately studies
multi-turn degradation, compaction loss, and error propagation from retained
experience
\citep{laban2026llms,li2026selfcompactinglanguagemodelagents,xiong2026memorymanagementimpactsllm}.
Our task orders are fixed by causal compatibility in \gaia{} and commit history
in \swe{}, which preserves joint feasibility without testing sensitivity to
ordering, and all tasks are independently specified rather than carrying
explicit cross-task dependencies. On the training side, we test Seq-RL on a
single 30B model, and our prefill setup spends most annotated data on context:
only 9.6K of the 27K scenarios receive gradient while the rest serve as frozen
prefills or go unused. Training on longer on-policy continuations without
frozen prefills would use more of this data and expose the policy to
self-generated long contexts, at higher rollout cost. Further tuning of the
reward, rollout length, and prefill strategy may yield more compute-efficient
recipes.

\paragraph{Exposing benchmark and infrastructure failures.}
Running agents without resets exposes flaws in benchmarks and infrastructure
that isolated evaluation hides, because a faulty tool, grader loophole, or
corrupted environment can compound across later tasks. Related work benchmarks
such shortcut and evaluator-exploitation behavior across isolated and
sequential tool-use settings \citep{thaman2026rewardhackingbenchmarkmeasuring}.
In an excluded pilot run with a misconfigured \swe{} container, sequential
agents learned from early tasks to inspect hidden commit history and retrieve
gold patches, and hit 100\% success by the third task. That exploit never
appeared under independent evaluation because each reset wiped out what the
agent had discovered.

\section{Conclusion}

We introduced sequential evaluation, a simple protocol that removes task-level
resets while keeping tasks and graders unchanged. Across \gaia{} and \swe{},
persistence produced model-dependent drops in accuracy and changed how every
model allocated effort, even when mean accuracy held steady. Agents reused
scripts and prior knowledge to shorten trajectories, but also carried unchecked
assumptions into later tasks. Training inside persistent sessions with Seq-RL
narrowed these sequential gaps and transferred across domains, harnesses, and
external memory benchmarks. Because sequential evaluation applies to any
benchmark whose tasks share an environment, it provides a direct testbed for
agents that learn their environment and users in context over time. Reliable
long-lived agents need more than single-task competence: they must manage their
accumulated experience.

\section*{Acknowledgements}
We thank Josh Gross and Nat Friedman for their leadership and support of this project. We also thank Tomas Hodan for his contributions during its initial stages, as well as the entire RL and Agents team at Meta Superintelligence Labs Paris.

\clearpage
\bibliographystyle{assets/plainnat}
\bibliography{paper}

\clearpage
\beginappendix

The appendix follows the structure of the main text: \autoref{app:adapting}
gives experimental details, \autoref{app:seq-eval-additional} reports
additional results for sequential evaluation (\autoref{sec:results}), and
\autoref{app:seqrl-additional} gives details and additional results for
Seq-RL (\autoref{sec:seq-rl}).

\section{Experimental details}
\label{app:adapting}

\subsection{Building the sequential benchmarks}
\label{app:benchmarks}

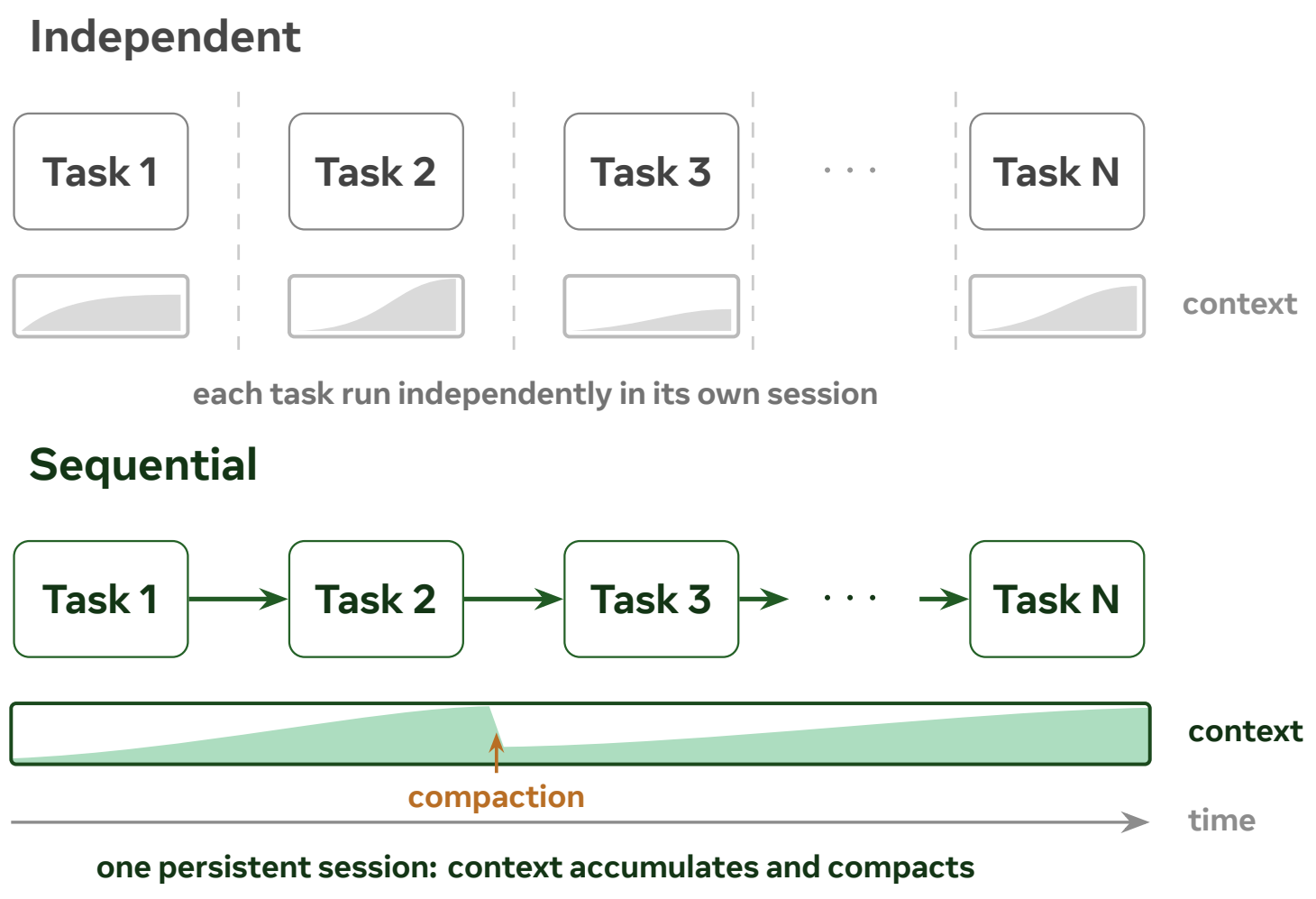
\begin{wrapfigure}{r}{0.58\linewidth}
\centering
\resizebox{\linewidth}{!}{%
\begin{tikzpicture}[
>={Stealth[length=2mm]},
task/.style={
rounded corners=3pt,
draw,
minimum width=12mm,
minimum height=8mm,
inner sep=1.5pt,
font=\sffamily\small\bfseries,
align=center
},
itask/.style={
task,
draw=indgray!95!black,
fill=white,
text=black!74
},
stask/.style={
task,
draw=seqgreen!78!black,
fill=white,
text=seqgreen!42!black
},
title/.style={font=\sffamily\bfseries},
clab/.style={font=\sffamily\scriptsize, text=black!45},
sub/.style={font=\sffamily\scriptsize, text=black!55},
sarr/.style={->, seqgreen!72!black, line width=0.9pt},
]
\def\xa{0}
\def\xb{1.9}
\def\xc{3.8}
\def\xN{6.6}
\def\xdots{5.2}
\def\xl{-0.62}
\def\xr{7.24}
\node[
title,
text=black!72,
anchor=west
] at (\xl,4.05) {Independent};
\foreach \x in {0.95,2.85,4.5,5.9}{
\draw[
indgray!45,
dashed,
line width=0.5pt
] (\x,3.7) -- (\x,1.92);
}
\foreach \lab/\x in {1/\xa,2/\xb,3/\xc,N/\xN}{
\node[itask] (I\lab) at (\x,3.15) {Task \lab};
}
\node[
font=\sffamily,
text=black!40
] at (\xdots,3.15) {$\cdots$};
\fill[indgray!32]
(-0.55,2.05)
.. controls (-0.28,2.28) and (0.12,2.30) ..
(0.55,2.30)
-- (0.55,2.05) -- cycle;
\fill[indgray!32]
(1.35,2.05)
.. controls (1.98,2.06) and (2.02,2.41) ..
(2.45,2.41)
-- (2.45,2.05) -- cycle;
\fill[indgray!32]
(3.25,2.05)
.. controls (3.80,2.09) and (3.95,2.20) ..
(4.35,2.20)
-- (4.35,2.05) -- cycle;
\fill[indgray!32]
(6.05,2.05)
.. controls (6.60,2.11) and (6.75,2.36) ..
(7.15,2.36)
-- (7.15,2.05) -- cycle;
\foreach \x in {0,1.9,3.8,6.6}{
\draw[
indgray!60,
line width=0.7pt,
rounded corners=1pt
] (\x-0.6,2.01) rectangle (\x+0.6,2.43);
}
\node[
clab,
anchor=west
] at (\xN+0.6+0.14,2.22) {context};
\node[sub] at (3.0,1.6)
{each task run independently in its own session};
\node[
title,
text=seqgreen!42!black,
anchor=west
] at (\xl,1.1) {Sequential};
\foreach \lab/\x in {1/\xa,2/\xb,3/\xc,N/\xN}{
\node[stask] (S\lab) at (\x,0.2) {Task \lab};
}
\node[
font=\sffamily,
text=seqgreen!45!black
] at (\xdots,0.2) {$\cdots$};
\draw[sarr] (S1) -- (S2);
\draw[sarr] (S2) -- (S3);
\draw[sarr] (S3) -- (4.75,0.2);
\draw[sarr] (5.65,0.2) -- (SN);
\begin{scope}
\clip (\xl,-0.94) rectangle (\xr,-0.52);
\fill[ctxgreen!45]
  (\xl,-0.94)
  -- (\xl,-0.90)
  .. controls (0.6,-0.85) and (1.9,-0.55) ..
  (2.68,-0.54)
  -- (2.78,-0.82)
  .. controls (4.2,-0.80) and (6.0,-0.57) ..
  (\xr,-0.55)
  -- (\xr,-0.94)
  -- cycle;
\end{scope}
\draw[
seqgreen!58!black,
line width=0.7pt,
rounded corners=1pt
] (\xl,-0.94) rectangle (\xr,-0.52);
\node[
font=\sffamily\scriptsize\bfseries,
text=seqorange!85!black,
anchor=north
] (compactlabel) at (2.73,-1.00) {compaction};
\draw[
-{Stealth[length=1.4mm]},
seqorange!85!black,
line width=0.65pt
] (compactlabel.north) -- (2.73,-0.72);
\node[
clab,
text=seqgreen!40!black,
anchor=west
] at (\xr+0.14,-0.73) {context};
\draw[
->,
black!45,
line width=0.6pt
] (\xl,-1.34) -- (\xr,-1.34);
\node[
clab,
anchor=west
] at (\xr+0.14,-1.34) {time};
\node[
sub,
text=seqgreen!42!black
] at (3.0,-1.67)
{one persistent session: context accumulates and compacts};
\end{tikzpicture}%
}
\vspace{4pt}
\caption{\textbf{Sequential evaluation setup.}
Independent evaluation runs each task in a fresh session. Sequential
evaluation processes independently specified tasks in one persistent
session, carrying conversation and files across tasks as context accumulates
and compacts. \gaia{} additionally preserves memory and environment state,
whereas \swe{} checks out each task's starting commit.}
\label{fig:eval-setup}
\end{wrapfigure}

\begin{figure}[tb]
  \centering
  \begin{minipage}[c]{0.46\linewidth}
    \centering
    \resizebox{\linewidth}{!}{%
  \begin{tikzpicture}[
    font=\sffamily,
    axis/.style={
      draw=black!42,
      line width=0.55pt
    },
    grid/.style={
      draw=black!11,
      densely dotted,
      line width=0.45pt
    },
    tick/.style={
      font=\sffamily\scriptsize,
      text=black!55
    },
    ylabel/.style={
      font=\sffamily\scriptsize\bfseries,
      text=black!68
    },
    charttitle/.style={
      font=\sffamily\bfseries\small
    },
  ]
  \path[use as bounding box] (-0.08,2.85) rectangle (6.18,7.15);

  \fill[
    seqblue!2,
    rounded corners=4pt
  ] (-0.08,3.48) rectangle (6.18,6.52);

  \draw[
    seqblue!28,
    rounded corners=4pt,
    line width=0.6pt
  ] (-0.08,3.48) rectangle (6.18,6.52);
  \def\gx{1.55}
  \def\gs{0.27}

  \node[
    charttitle,
    text=seqblue!82!black,
    anchor=west
  ] at (0.15,6.23) {(a) \gaia{}};

  \node[ylabel,anchor=east] at (1.42,5.52) {Search};
  \node[ylabel,anchor=east] at (1.42,4.90) {Execution};

  \foreach \v in {5,10,15,20}{
    \pgfmathsetmacro{\xx}{\gx+(\v-4)*\gs}

    \draw[grid] (\xx,4.43) -- (\xx,5.82);
    \draw[axis] (\xx,4.43) -- ++(0,-0.07);

    \node[
      tick,
      anchor=north
    ] at (\xx,4.35) {\v};
  }

  \draw[axis]
    (\gx,4.43) -- ({\gx+16*\gs},4.43);

  \foreach \v/\dy in {
    11/0.02,
    12/-0.05,
    15/0.08,
    15/-0.08,
    16/0.00,
    17/0.09,
    18/-0.10,
    18/0.03,
    18/0.13,
    20/0.01
  }{
    \fill[seqblue!78]
      ({\gx+(\v-4)*\gs},{5.52+\dy})
      circle (1.7pt);
  }

  \foreach \v/\dy in {
    4/0.02,
    7/-0.04,
    8/0.06,
    8/-0.07,
    9/0.01,
    10/-0.04,
    12/0.05,
    12/-0.07,
    12/0.00,
    13/-0.10
  }{
    \fill[seqorange!90]
      ({\gx+(\v-4)*\gs},{4.90+\dy})
      circle (1.7pt);
  }

  \draw[
    seqblue,
    line width=1.6pt
  ] ({\gx+(16-4)*\gs},5.30)
    -- ({\gx+(16-4)*\gs},5.74);

  \draw[
    seqorange,
    line width=1.6pt
  ] ({\gx+(9.5-4)*\gs},4.68)
    -- ({\gx+(9.5-4)*\gs},5.12);

  \node[
    tick,
    text=black!65
  ] at ({\gx+8*\gs},3.68)
    {Scenarios per sequence};
  \end{tikzpicture}%
    }
  \end{minipage}\hfill
  \begin{minipage}[c]{0.50\linewidth}
    \centering
    \resizebox{\linewidth}{!}{%
  \begin{tikzpicture}[
    font=\sffamily,
    axis/.style={
      draw=black!42,
      line width=0.55pt
    },
    grid/.style={
      draw=black!11,
      densely dotted,
      line width=0.45pt
    },
    tick/.style={
      font=\sffamily\scriptsize,
      text=black!55
    },
    ylabel/.style={
      font=\sffamily\scriptsize\bfseries,
      text=black!68
    },
    charttitle/.style={
      font=\sffamily\bfseries\small
    },
  ]
  \fill[
    seqgreen!2,
    rounded corners=4pt
  ] (-0.08,-1.10) rectangle (6.18,3.20);

  \draw[
    seqgreen!28,
    rounded corners=4pt,
    line width=0.6pt
  ] (-0.08,-1.10) rectangle (6.18,3.20);
  \def\bx{1.95}
  \def\bs{0.0385}

  \begin{scope}[yshift=-0.40cm]

  \node[
    charttitle,
    text=seqgreen!55!black,
    anchor=west
  ] at (0.15,3.31) {(b) \swe{}};

  \foreach \v in {0,25,50,75,100}{
    \pgfmathsetmacro{\xx}{\bx+\v*\bs}

    \draw[grid] (\xx,0.10) -- (\xx,3.08);
    \draw[axis] (\xx,0.10) -- ++(0,-0.07);

    \node[
      tick,
      anchor=north
    ] at (\xx,0.01) {\v};
  }

  \draw[axis]
    (\bx,0.10) -- ({\bx+100*\bs},0.10);

  \foreach \repo/\value/\yy in {
    tutanota/20/2.90,
    NodeBB/44/2.64,
    element-web/56/2.38,
    navidrome/57/2.12,
    vuls/62/1.86,
    webclients/65/1.60,
    teleport/76/1.34,
    qutebrowser/79/1.08,
    flipt/85/0.82,
    openlibrary/91/0.56,
    ansible/96/0.30
  }{
    \pgfmathsetmacro{\xfull}{\bx+\value*\bs}
    \pgfmathsetmacro{\used}{min(\value,40)}
    \pgfmathsetmacro{\xused}{\bx+\used*\bs}

    \node[
      ylabel,
      anchor=east
    ] at ({\bx-0.10},\yy) {\repo};

    \draw[
      indgray!28,
      line width=4.4pt,
      line cap=round
    ] (\bx,\yy) -- (\xfull,\yy);

    \draw[
      seqgreen!72!black,
      line width=4.4pt,
      line cap=round
    ] (\bx,\yy) -- (\xused,\yy);

    \node[
      tick,
      text=black!45,
      anchor=west
    ] at ({\xfull+0.08},\yy) {\value};
  }

  \pgfmathsetmacro{\capx}{\bx+40*\bs}

  \draw[
    black!45,
    densely dashed,
    line width=0.65pt
  ] (\capx,0.10) -- (\capx,3.10);

  \node[
    font=\sffamily\scriptsize\bfseries,
    text=black!60,
    anchor=south west
  ] at ({\capx+0.08},3.05) {cap 40};

  \node[
    tick,
    text=black!65
  ] at ({\bx+50*\bs},-0.50)
    {Issues per repository};

  \end{scope}
  \end{tikzpicture}%
    }
  \end{minipage}
  \caption{\textbf{Sequence lengths.} \emph{(a)}~\gaia{} scenarios per
  sequence for the Search and Execution splits: 255 tasks in 20 sequences
  of 4--20. \emph{(b)}~\swe{} issues per repository, where green marks the tasks
  retained in each sequence under the 40-task cap. The cap binds for ten of
  the eleven repositories, while \texttt{tutao/tutanota} contributes only
  20, giving 420 tasks in total.}
  \label{fig:seq-lengths}
\end{figure}
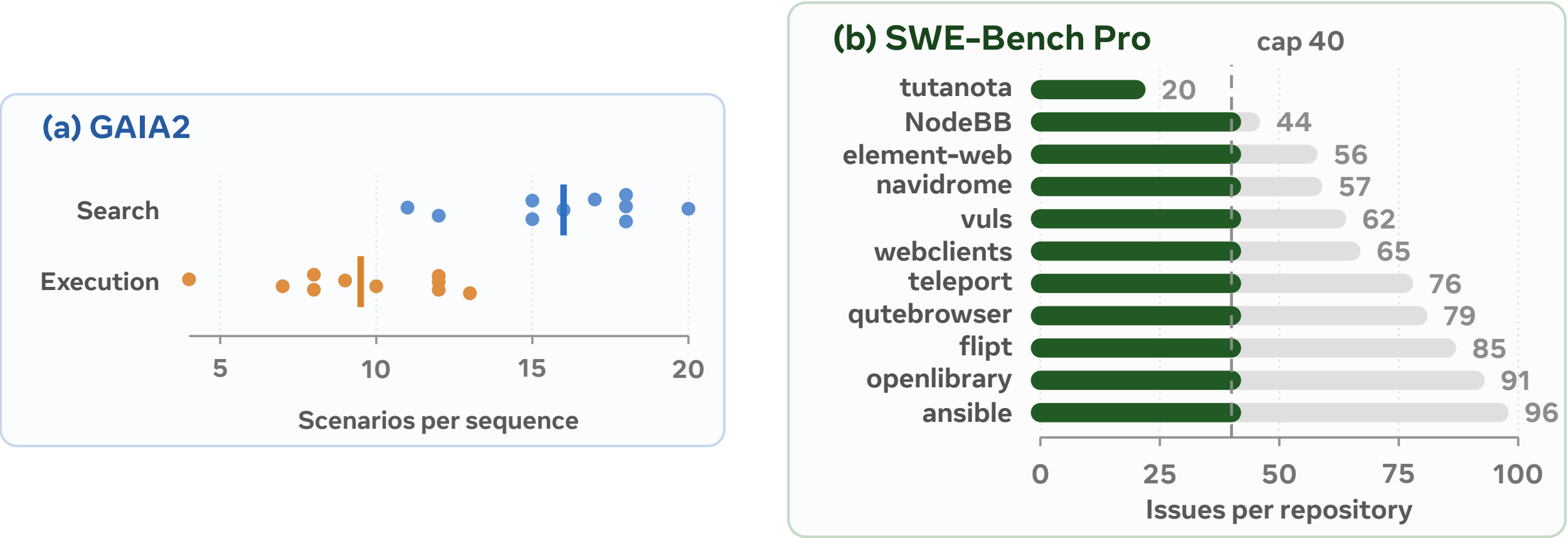

We convert \gaia{} and \swe{} into sequential benchmarks by removing
task-level session resets while keeping task definitions and grading
unchanged. Tasks are grouped by their shared environment (a user universe in
\gaia{}, a Git repository in \swe{}), and each group forms one persistent
sequence. \autoref{fig:eval-setup} illustrates the protocol and
\autoref{fig:seq-lengths} the resulting sequence lengths.

\paragraph{\gaia{}.}
Each universe and split defines one sequence. In the original benchmark,
every scenario starts from the same initial universe state. Sequentially,
scenario $t+1$ starts from the state left by scenario $t$. We therefore order
scenarios so that completing one as intended never makes a later one
infeasible. A deterministic checker uses \gaia{}'s write-action DAG
annotations to detect conflicting state modifications, and a Claude Code
review flags semantic conflicts that the annotations miss. Of the 160
Execution scenarios, only 95 can be arranged into compatible sequences, so we drop the remaining
65 and keep all 160 Search scenarios. Compatibility
holds under the intended solutions only: the effects of erroneous agent
actions are deliberately allowed to propagate. Neither the environment nor
the OpenClaw session is reset, so conversation, memory, files, compactions,
and environment changes persist throughout each sequence. Independent
evaluation uses the same 255 scenarios, so every comparison is paired.

\paragraph{\swe{}.}
Each repository defines one sequence. We order issues by their starting
commits, so that tasks follow the evolution of the codebase, and keep the
first 40 per repository for 420 tasks in 11 sequences. Before each
task, we clean the working tree, check out the task's starting commit, and
notify the agent, so that it does not assume its previous patch is still
applied. Agent-visible checkouts are scrubbed of Git history and artifacts
that could reveal reference patches, and the agent has no internet access. In principle, the agent can still inspect how the repository evolved after an
earlier issue, though we did not observe agents recovering reference solutions
this way. Only the repository checkout changes between tasks:
conversation, compaction history, and the agent's home directory persist, as
for a coding agent that moves between checkouts without restarting its
session.

\subsection{Harnesses}
\label{app:mini-swe}

\paragraph{OpenClaw (\gaia{}).}
We use OpenClaw's native context management, which has two mechanisms:
tool-output compaction, which compacts old tool outputs, and full-history
summarization, which summarizes older history when the context nears the
window limit. \autoref{tab:compaction} reports how often each occurs.

\paragraph{Mini-SWE-Agent (\swe{}).}
Mini-SWE-Agent~\citep{yang2024sweagent} is designed for independent,
single-task episodes: each task runs in a fresh container, and
\texttt{DefaultAgent.run()} resets the conversation. We add three mechanisms
for sequential evaluation. Disabling them recovers the stock behavior, so
both modes run the same agent.
\begin{itemize}
  \item \textbf{Persistent conversation session.} A session object keeps the
  message history across tasks, so task $t{+}1$ continues the conversation of
  task $t$.
  \item \textbf{Context compaction.} When the conversation nears the window
  limit, older history is condensed into a structured summary of goals,
  progress, and next steps while recent turns are kept, mirroring OpenClaw's
  full-history summarization.
  \item \textbf{Agent-home persistence.} The agent's home directory persists
  across tasks, while the repository switches to each task's starting commit.
\end{itemize}
The agent has no dedicated cross-task memory module: beyond the conversation,
persistence is limited to files it creates in its home directory.

\subsection{Evaluation protocol and statistics}
\label{app:eval-stats}

We run $K=5$ rollouts per \gaia{} scenario and $K=3$ per \swe{} task in each
mode. Each sequential rollout runs a complete sequence from a clean state.
Infrastructure failures, such as VM failures, lost connectivity, or SSH tunnel
and podman exceptions, are replayed so that they do not propagate to later
tasks in the sequence.

\paragraph{Metrics.}
$\mathrm{avg@}K$ is the mean success rate over the $K$ rollouts,
$\mathrm{pass@}K$ the fraction of tasks solved in at least one rollout, and
$\mathrm{pass}^{K}$ the fraction solved in all $K$. We report differences as
$\Delta=\mathrm{seq}-\mathrm{ind}$, so negative values indicate degradation.
We use two Python measures: \emph{Python share}, the fraction of tool calls
that execute Python (\autoref{tab:python-usage}), and \emph{Python use}, the
fraction of scenario runs that call Python at least once
(\autoref{fig:behavioral-adaptation}(a), \autoref{fig:seqrl-behavior-position}(a),
and \autoref{tab:seqrl-gaia2-behavior}).

\paragraph{Statistics.}
We report 95\% cluster-bootstrap confidence intervals for $\mathrm{avg@}K$
only, while $\mathrm{pass@}K$ and $\mathrm{pass}^{K}$ are point estimates. The
resampling unit is the task for independent results, the complete sequence
for sequential results, and the matched sequence for $\Delta$. For
\autoref{app:ablations}, we compute Spearman correlations between the
number of tool-output compactions in a \gaia{} scenario and its outcome
change from independent to sequential, coded as regression ($-1$), stable
($0$), or recovery ($+1$).

\FloatBarrier
\section{Sequential evaluation: additional results}
\label{app:seq-eval-additional}
\suppressfloats[t]

This section reports the complete results behind \autoref{sec:results}.
Results use a 200K-token window unless stated otherwise.

\subsection{Task performance}
\label{app:task-performance}

\autoref{tab:abs-gaia2} and \autoref{tab:abs-swe} report full $\mathrm{avg@}K$
(with 95\% CIs), $\mathrm{pass@}K$, and $\mathrm{pass}^{K}$ results for both
modes. Because $\mathrm{pass@}K$ stays close to its independent value in every
setting, the $\mathrm{avg@}K$ drops in \autoref{sec:results-accuracy} come from
tasks solved in fewer rollouts rather than tasks no longer solved at all.

\begin{table}[!ht]
\centering
\caption{\textbf{\gaia{} full metrics report at 200K ($K=5$).} Values are percentages with $\Delta=\mathrm{Seq.}-\mathrm{Indep.}$ For $\mathrm{avg@}5$, $\pm$ gives the 95\% cluster-bootstrap half-width in gray (\autoref{app:eval-stats}), and other metrics are point estimates. Bold marks the best value per metric and split.}
\label{tab:abs-gaia2}
\scriptsize
\setlength{\tabcolsep}{2.5pt}
\renewcommand{\arraystretch}{1.03}
\newcommand{\bestscore}[1]{\boldsymbol{#1}}
\newcommand{\scoreci}[2]{#1{\color{black!50}\scriptstyle\,\pm#2}}

\begin{tabularx}{\linewidth}{@{}ll*{6}{>{\raggedleft\arraybackslash}X}@{}}
\toprule
& & \multicolumn{3}{c}{\textbf{\gaia{} Search}}
& \multicolumn{3}{c}{\textbf{\gaia{} Execution}} \\
\cmidrule(lr){3-5}\cmidrule(lr){6-8}
\textbf{Metric} & \textbf{Model}
& \textbf{Indep.} & \textbf{Seq.} & $\Delta$
& \textbf{Indep.} & \textbf{Seq.} & $\Delta$ \\
\midrule
\multirow{3}{*}{avg@5}
& \opus{}
& $\scoreci{87.6}{3.8}$
& $\scoreci{86.2}{6.0}$
& $\scoreci{-1.4}{4.6}$
& $\scoreci{\bestscore{83.2}}{6.2}$
& $\scoreci{\bestscore{80.2}}{6.1}$
& $\scoreci{-3.0}{5.3}$ \\

& \gpt{}
& $\scoreci{\bestscore{90.2}}{3.4}$
& $\scoreci{\bestscore{92.1}}{1.9}$
& $\scoreci{\bestscore{+1.9}}{3.0}$
& $\scoreci{70.9}{7.1}$
& $\scoreci{68.2}{8.6}$
& $\scoreci{\bestscore{-2.7}}{4.3}$ \\

& \dps{}
& $\scoreci{81.3}{4.3}$
& $\scoreci{76.0}{2.4}$
& $\scoreci{-5.3}{3.8}$
& $\scoreci{70.1}{7.0}$
& $\scoreci{63.8}{5.6}$
& $\scoreci{-6.3}{7.5}$ \\
\midrule
\multirow{3}{*}{pass@5}
& \opus{}
& $96.2$ & $\bestscore{98.8}$ & $\bestscore{+2.6}$
& $\bestscore{92.6}$ & $\bestscore{90.5}$ & $-2.1$ \\

& \gpt{}
& $\bestscore{98.1}$ & $96.9$ & $-1.2$
& $89.5$ & $89.5$ & $\bestscore{0.0}$ \\

& \dps{}
& $96.2$ & $96.2$ & $0.0$
& $87.4$ & $86.3$ & $-1.1$ \\
\midrule
\multirow{3}{*}{pass$^5$}
& \opus{}
& $70.0$ & $68.8$ & $-1.2$
& $\bestscore{67.4}$ & $\bestscore{61.1}$ & $\bestscore{-6.3}$ \\

& \gpt{}
& $\bestscore{76.2}$ & $\bestscore{81.2}$ & $\bestscore{+5.0}$
& $47.4$ & $33.7$ & $-13.7$ \\

& \dps{}
& $58.8$ & $42.5$ & $-16.3$
& $38.9$ & $30.5$ & $-8.4$ \\
\bottomrule
\end{tabularx}
\end{table}

\begin{table}[!ht]
\centering
\caption{\textbf{\swe{} full metrics report at 200K ($K=3$).} Values are percentages with $\Delta=\mathrm{Seq.}-\mathrm{Indep.}$ For $\mathrm{avg@}3$, $\pm$ gives the 95\% cluster-bootstrap half-width in gray (\autoref{app:eval-stats}), and other metrics are point estimates. Bold marks the best value per metric.}
\label{tab:abs-swe}
\scriptsize
\setlength{\tabcolsep}{7pt}
\renewcommand{\arraystretch}{1.03}

\newcommand{\bestscore}[1]{\boldsymbol{#1}}
\newcommand{\scoreci}[2]{#1{\color{black!50}\scriptstyle\,\pm#2}}

\begin{tabularx}{0.72\linewidth}
{@{}ll*{3}{>{\raggedleft\arraybackslash}X}@{}}
\toprule
\textbf{Metric} & \textbf{Model}
& \textbf{Indep.} & \textbf{Seq.} & $\Delta$ \\
\midrule

\multirow{3}{*}{avg@3}
& \opus{}
& $\scoreci{51.0}{3.5}$
& $\scoreci{50.2}{3.4}$
& $\scoreci{-0.8}{0.9}$ \\

& \gpt{}
& $\scoreci{\bestscore{53.8}}{3.9}$
& $\scoreci{\bestscore{53.6}}{3.7}$
& $\scoreci{\bestscore{-0.2}}{1.2}$ \\

& \dps{}
& $\scoreci{50.4}{3.5}$
& $\scoreci{44.3}{3.1}$
& $\scoreci{-6.1}{2.7}$ \\

\midrule
\multirow{3}{*}{pass@3}
& \opus{}
& $56.2$ & $57.4$ & $\bestscore{+1.2}$ \\

& \gpt{}
& $\bestscore{60.2}$ & $\bestscore{60.2}$ & $0.0$ \\

& \dps{}
& $59.0$ & $56.4$ & $-2.6$ \\

\midrule
\multirow{3}{*}{pass$^3$}
& \opus{}
& $44.0$ & $\bestscore{40.0}$ & $\bestscore{-4.0}$ \\

& \gpt{}
& $\bestscore{44.5}$ & $36.9$ & $-7.6$ \\

& \dps{}
& $41.2$ & $30.5$ & $-10.7$ \\
\bottomrule
\end{tabularx}
\end{table}

\subsection{Behavioral changes}
\label{app:behavior}
\label{app:gaia2-eff}
\label{app:gaia2-scripts}
\label{app:swe-actions}

\begin{figure}[!ht]
  \centering
  \includegraphics[width=\columnwidth]{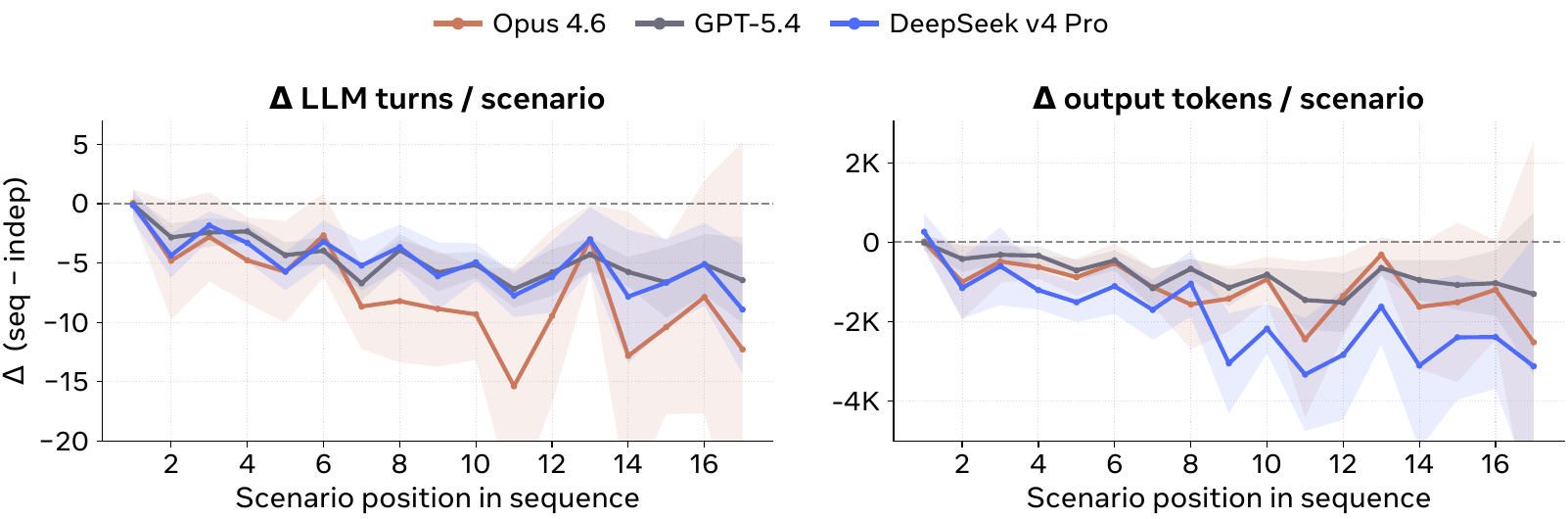}
  \caption{\textbf{\gaia{} effort by sequence position.} Change from independent to sequential ($\Delta=\mathrm{seq}-\mathrm{indep}$) in LLM turns (left) and output tokens (right) per scenario, where negative values indicate less effort sequentially. Bands show 95\% confidence intervals.}
  \label{fig:eff-gaia2}
\end{figure}

On \gaia{}, sequential runs already use fewer turns and output tokens at the
second scenario of a sequence (\autoref{fig:eff-gaia2}), long before the
context approaches the window limit (\autoref{app:ablations}), so the
reduction reflects a response to prior context rather than context pressure.
Python's share of tool calls rises with persistence, also rises when only the
window shrinks, and is highest when both apply (\autoref{tab:python-usage}):
agents increasingly process tool output in scripts that keep raw records out
of the context, without prompting (\autoref{fig:gaia2-scripts}).
Because these scripts remain available to later tasks, their assumptions carry
forward as well. Among 1,574 sequential scenario runs in which Python
automates pagination, scripts whose termination logic assumes a fixed bound
succeed in 77.2\% of runs, against 86.7\% for scripts that verify complete
retrieval (pooled over models and tasks, unadjusted, and observational).

\begin{table}[!ht]
  \centering
  \scriptsize
  \setlength{\tabcolsep}{7pt}
  \renewcommand{\arraystretch}{1.12}
  \caption{\textbf{Python share on \gaia{}.} Fraction of tool calls that execute Python, with Search and Execution pooled.}
  \label{tab:python-usage}
  \begin{tabular}{@{}l@{\hspace{12pt}}rr@{\hspace{12pt}}rr@{}}
    \toprule
    & \multicolumn{2}{c}{\textbf{200K}} & \multicolumn{2}{c}{\textbf{32K}} \\
    \cmidrule(lr){2-3}\cmidrule(lr){4-5}
    \textbf{Model} & \textbf{Indep.} & \textbf{Seq.} & \textbf{Indep.} & \textbf{Seq.} \\
    \midrule
    \opus{}        & 2.1\% &  8.9\% &  9.5\% & 19.1\% \\
    \gpt{}         & 9.9\% & 15.3\% & 13.2\% & 24.2\% \\
    \dps{} & 2.6\% & 10.9\% &  6.3\% & 16.6\% \\
    \bottomrule
  \end{tabular}
\end{table}

\begin{figure}[!ht]
\begin{promptbox}{(a) Fixed-bound helper script in {\mdseries\texttt{/home/agent/process\_data.py}}}
import json, subprocess
def get_contacts(offset):
    r = subprocess.run(["contacts", "get-contacts", "--offset", str(offset)],
                       capture_output=True, text=True)
    return json.loads(r.stdout)

contacts = []
for off in range(0, 130, 10):
    contacts.extend(get_contacts(off)["contacts"])
# Assumes that 130 records are sufficient.
\end{promptbox}

\vspace{4pt}

\begin{promptbox}{(b) Completeness-checked {\mdseries\texttt{search\_contacts}} helper}
all_contacts, offset = [], 0
while True:
    data = get_contacts(offset)
    page = data["contacts"]
    all_contacts.extend(page)
    if offset + len(page) >= data["metadata"]["total"]:
        break
    offset += len(page)
# Re-implement the unsupported compound query
matches = [c for c in all_contacts
           if c["nationality"] == "Swedish" and c["age"] < 40 and not c["is_user"]]
\end{promptbox}
\caption{\textbf{Persistent home-directory scripts on \gaia{} Search.}
Condensed excerpts contrast fixed-bound pagination (a) with an \opus{}
agent's completeness-checked pagination and local implementation of a compound
query unsupported by \texttt{search\_contacts} (b). Both keep raw records out
of the context window, so correctness depends on the scripts' termination
assumptions.}
\label{fig:gaia2-scripts}
\end{figure}

\autoref{tab:swe-actions} reports the full \swe{} workflow breakdown behind
\autoref{sec:behavior}: across all three frontier models, shorter sequential
trajectories come primarily from inspecting fewer files and reaching the first
edit sooner, while the share of turns spent testing rises.

\begin{table}[!ht]
  \centering
  \scriptsize
  \setlength{\tabcolsep}{4.5pt}
  \renewcommand{\arraystretch}{1.12}
  \caption{\textbf{\swe{} effort and workflow at 200K.} Means per task, pooled over the 420 tasks of each run, with relative changes from independent to sequential in gray. \emph{To 1st edit}: turns before the first file edit. \emph{Files}: distinct source files referenced. \emph{Inspect/Edit/Test}: share of turns whose dominant action is inspection, editing, or testing.}
  \label{tab:swe-actions}
  \newcommand{\rel}[1]{\,{\color{black!50}\ensuremath{\scriptstyle(-#1\%)}}}
  \newcommand{\norel}{\phantom{\rel{00}}}
  \begin{tabular}{@{}ll@{\hspace{14pt}}rr@{\hspace{14pt}}rr@{\hspace{14pt}}rrr@{}}
    \toprule
    & & \multicolumn{2}{c}{\textbf{Effort}}
      & \multicolumn{2}{c}{\textbf{Exploration}}
      & \multicolumn{3}{c}{\textbf{Share of turns (\%)}} \\
    \cmidrule(lr){3-4}\cmidrule(lr){5-6}\cmidrule(lr){7-9}
    \textbf{Model} & \textbf{Mode} & \textbf{Turns}
      & \shortstack{\textbf{Output}\\\textbf{tokens}}
      & \shortstack{\textbf{To 1st}\\\textbf{edit}}
      & \textbf{Files} & \textbf{Inspect} & \textbf{Edit} & \textbf{Test} \\
    \midrule
    \multirow{2}{*}{\opus{}}
      & Indep. & 49.3\norel & 14.6K\norel & 19.7 & 15.9 & 71.5 & 11.1 & 8.6 \\
      & Seq.   & 28.2\rel{43} & 7.9K\rel{46}
               & 12.6 & 11.2 & 67.4 & 10.1 & 10.3 \\
    \midrule
    \multirow{2}{*}{\gpt{}}
      & Indep. & 24.1\norel & 31.3K\norel & 10.4 & 16.0 & 57.0 & 15.0 & 7.3 \\
      & Seq.   & 17.8\rel{26} & 25.1K\rel{20}
               & 8.6 & 12.1 & 55.7 & 9.9 & 11.4 \\
    \midrule
    \multirow{2}{*}{\dps{}}
      & Indep. & 57.9\norel & 31.1K\norel & 22.0 & 14.8 & 72.5 & 10.8 & 3.5 \\
      & Seq.   & 28.3\rel{51} & 12.6K\rel{60}
               & 13.0 & 10.1 & 64.8 & 6.0 & 7.3 \\
    \bottomrule
  \end{tabular}
\end{table}

\subsection{Context pressure and window size}
\label{app:ablations}
\label{app:compaction}
\label{app:gaia2-32k}
\label{app:swe-1m}

\begin{figure}[!ht]
  \centering
  \includegraphics[width=0.83\columnwidth]{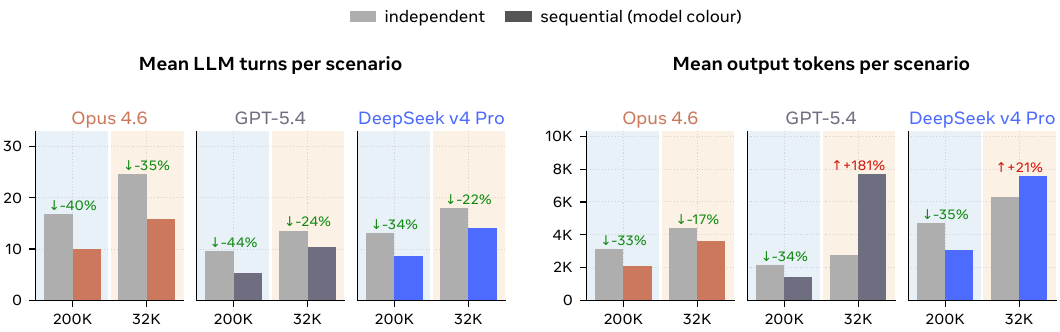}
  \caption{\textbf{\gaia{} effort at 200K and 32K.} Mean LLM turns (left) and output tokens (right) per scenario, with Search and Execution pooled. Labels give the relative change from independent to sequential.}
  \label{fig:eff-32k}
\end{figure}

We test whether the window size explains the effects of persistence by
shrinking it to 32K on \gaia{} and expanding it to 1M on \swe{}. On \gaia{} at
200K, OpenClaw never summarizes the full history in either mode, while
sequential runs compact old tool outputs an order of magnitude more often than
independent runs (\autoref{tab:compaction}): the sequential effects arise
without the context filling up. Coding independent-to-sequential outcome
changes as regressions ($-1$), stable outcomes ($0$), or recoveries ($+1$), we
find that tool-output compaction count is negatively associated with outcome
change for all models, and significantly so for \gpt{} ($\rho=-0.140$,
$p<0.001$) and \dps{} ($\rho=-0.072$, $p=0.012$). For \gpt{}, regressions average
33.6 compactions, compared with 15.0 for stable outcomes and 10.7 for
recoveries. These correlations do not establish that compaction causes
regressions: difficult trajectories may generate more tool output that causes both
more compactions and worse outcomes. We also find no clear relationship
between a task's position in the sequence and its performance change. On \swe{},
compactions are not correlated with performance changes, possibly because each issue starts from its own commit and successive issues
often touch different parts of the codebase. At 32K, independent runs also summarize and
lose accuracy (\autoref{tab:abs-gaia2-32k}), so the comparison is between two
constrained settings, and the sequential gap remains model-dependent.
Sequential runs still take fewer turns, but two of the three models now
produce more output tokens (\autoref{fig:eff-32k}) as they regenerate
context lost to summarization.

\begin{table}[!ht]
  \centering
  \scriptsize
  \setlength{\tabcolsep}{4.5pt}
  \renewcommand{\arraystretch}{1.12}
  \caption{\textbf{Compaction on \gaia{}.} Average compaction events per scenario, by model, split, context window, and evaluation mode.}
  \label{tab:compaction}
  \begin{tabular}{@{}ll@{\hspace{14pt}}rr@{\hspace{8pt}}rr@{\hspace{16pt}}rr@{\hspace{8pt}}rr@{}}
    \toprule
    & &
    \multicolumn{4}{c}{\textbf{(a) Tool-output compactions}} &
    \multicolumn{4}{c}{\textbf{(b) Full-history summarizations}} \\
    \cmidrule(lr){3-6}\cmidrule(lr){7-10}
    & &
    \multicolumn{2}{c}{\textbf{200K}} &
    \multicolumn{2}{c}{\textbf{32K}} &
    \multicolumn{2}{c}{\textbf{200K}} &
    \multicolumn{2}{c}{\textbf{32K}} \\
    \cmidrule(lr){3-4}\cmidrule(lr){5-6}
    \cmidrule(lr){7-8}\cmidrule(lr){9-10}
    \textbf{Model} & \textbf{Split}
      & \textbf{Indep.} & \textbf{Seq.} & \textbf{Indep.} & \textbf{Seq.}
      & \textbf{Indep.} & \textbf{Seq.} & \textbf{Indep.} & \textbf{Seq.} \\
    \midrule
    \multirow{2}{*}{\opus{}} & Search
      & 3.6 & 47.7 & 33.7 & 37.9
      & 0 & 0 & 0.37 & 1.60 \\
    & Execution
      & 1.3 & 22.0 & 14.7 & 53.6
      & 0 & 0 & 0.55 & 1.83 \\
    \midrule
    \multirow{2}{*}{\gpt{}} & Search
      & 0.4 & 10.8 & 8.7 & 18.8
      & 0 & 0 & 0.54 & 1.56 \\
    & Execution
      & 0.3 & 25.2 & 12.4 & 35.3
      & 0 & 0 & 0.51 & 1.58 \\
    \midrule
    \multirow{2}{*}{\dps{}} & Search
      & 2.2 & 38.7 & 19.0 & 25.2
      & 0 & 0 & 0.44 & 1.81 \\
    & Execution
      & 0.6 & 24.1 & 14.6 & 41.9
      & 0 & 0 & 0.77 & 2.04 \\
    \bottomrule
  \end{tabular}
\end{table}

\begin{table}[!ht]
\centering
\caption{\textbf{\gaia{} full metrics report at 32K ($K=5$).} Values are percentages with $\Delta=\mathrm{Seq.}-\mathrm{Indep.}$ For $\mathrm{avg@}5$, $\pm$ gives the 95\% cluster-bootstrap half-width in gray (\autoref{app:eval-stats}), and other metrics are point estimates. Bold marks the best value per metric and split.}
\label{tab:abs-gaia2-32k}
\scriptsize
\setlength{\tabcolsep}{2.5pt}
\renewcommand{\arraystretch}{1.03}
\newcommand{\bestscore}[1]{\boldsymbol{#1}}
\newcommand{\scoreci}[2]{#1{\color{black!50}\scriptstyle\,\pm#2}}

\begin{tabularx}{\linewidth}{@{}ll*{6}{>{\raggedleft\arraybackslash}X}@{}}
\toprule
& & \multicolumn{3}{c}{\textbf{\gaia{} Search}}
& \multicolumn{3}{c}{\textbf{\gaia{} Execution}} \\
\cmidrule(lr){3-5}\cmidrule(lr){6-8}
\textbf{Metric} & \textbf{Model}
& \textbf{Indep.} & \textbf{Seq.} & $\Delta$
& \textbf{Indep.} & \textbf{Seq.} & $\Delta$ \\
\midrule
\multirow{3}{*}{avg@5}
& \opus{}
& $\scoreci{82.8}{4.5}$
& $\scoreci{\bestscore{88.4}}{3.8}$
& $\scoreci{\bestscore{+5.6}}{3.2}$
& $\scoreci{\bestscore{75.4}}{7.0}$
& $\scoreci{\bestscore{73.6}}{4.9}$
& $\scoreci{-1.8}{4.6}$ \\

& \gpt{}
& $\scoreci{\bestscore{87.9}}{3.6}$
& $\scoreci{84.4}{4.2}$
& $\scoreci{-3.5}{4.9}$
& $\scoreci{67.6}{7.0}$
& $\scoreci{68.6}{7.2}$
& $\scoreci{\bestscore{+1.0}}{6.3}$ \\

& \dps{}
& $\scoreci{77.0}{4.5}$
& $\scoreci{75.3}{4.3}$
& $\scoreci{-1.7}{5.5}$
& $\scoreci{58.4}{7.1}$
& $\scoreci{56.7}{5.0}$
& $\scoreci{-1.7}{6.1}$ \\
\midrule
\multirow{3}{*}{pass@5}
& \opus{}
& $95.0$ & $\bestscore{98.8}$ & $\bestscore{+3.8}$
& $\bestscore{90.5}$ & $\bestscore{92.6}$ & $\bestscore{+2.1}$ \\

& \gpt{}
& $\bestscore{97.5}$ & $97.5$ & $0.0$
& $\bestscore{90.5}$ & $90.5$ & $0.0$ \\

& \dps{}
& $95.0$ & $98.1$ & $+3.1$
& $89.5$ & $85.3$ & $-4.2$ \\
\midrule
\multirow{3}{*}{pass$^5$}
& \opus{}
& $64.4$ & $\bestscore{71.9}$ & $\bestscore{+7.5}$
& $\bestscore{57.9}$ & $\bestscore{43.2}$ & $-14.7$ \\

& \gpt{}
& $\bestscore{68.1}$ & $58.1$ & $-10.0$
& $37.9$ & $30.5$ & $\bestscore{-7.4}$ \\

& \dps{}
& $42.5$ & $40.0$ & $-2.5$
& $30.5$ & $22.1$ & $-8.4$ \\
\bottomrule
\end{tabularx}
\end{table}

\begin{figure}[!ht]
  \centering
  \includegraphics[width=0.72\columnwidth]{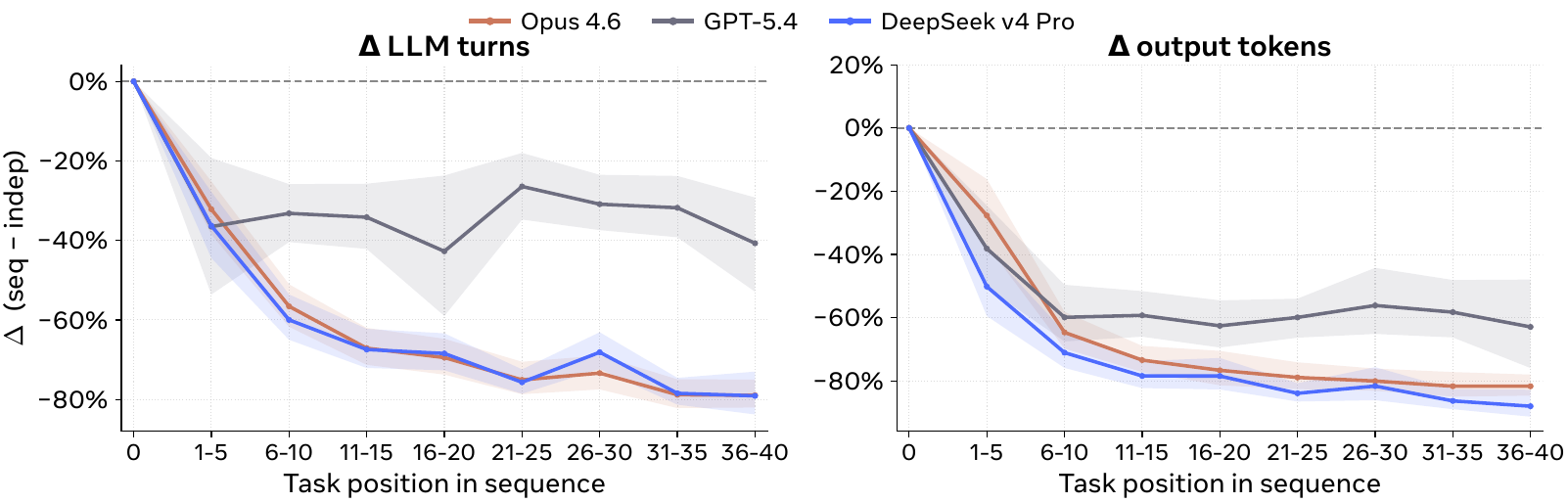}
  \caption{\textbf{\swe{} effort reduction at a 1M context window by task position.} Change from independent to sequential evaluation at 1M in LLM turns (left) and output tokens (right), with 95\% confidence intervals. Aggregate $\mathrm{avg@}3$, turns, and output tokens appear in \autoref{fig:swe-1m-main}.}
  \label{fig:swe-1m}
\end{figure}

\begin{figure}[p]
  \centering
  \includegraphics[width=0.9\textwidth]{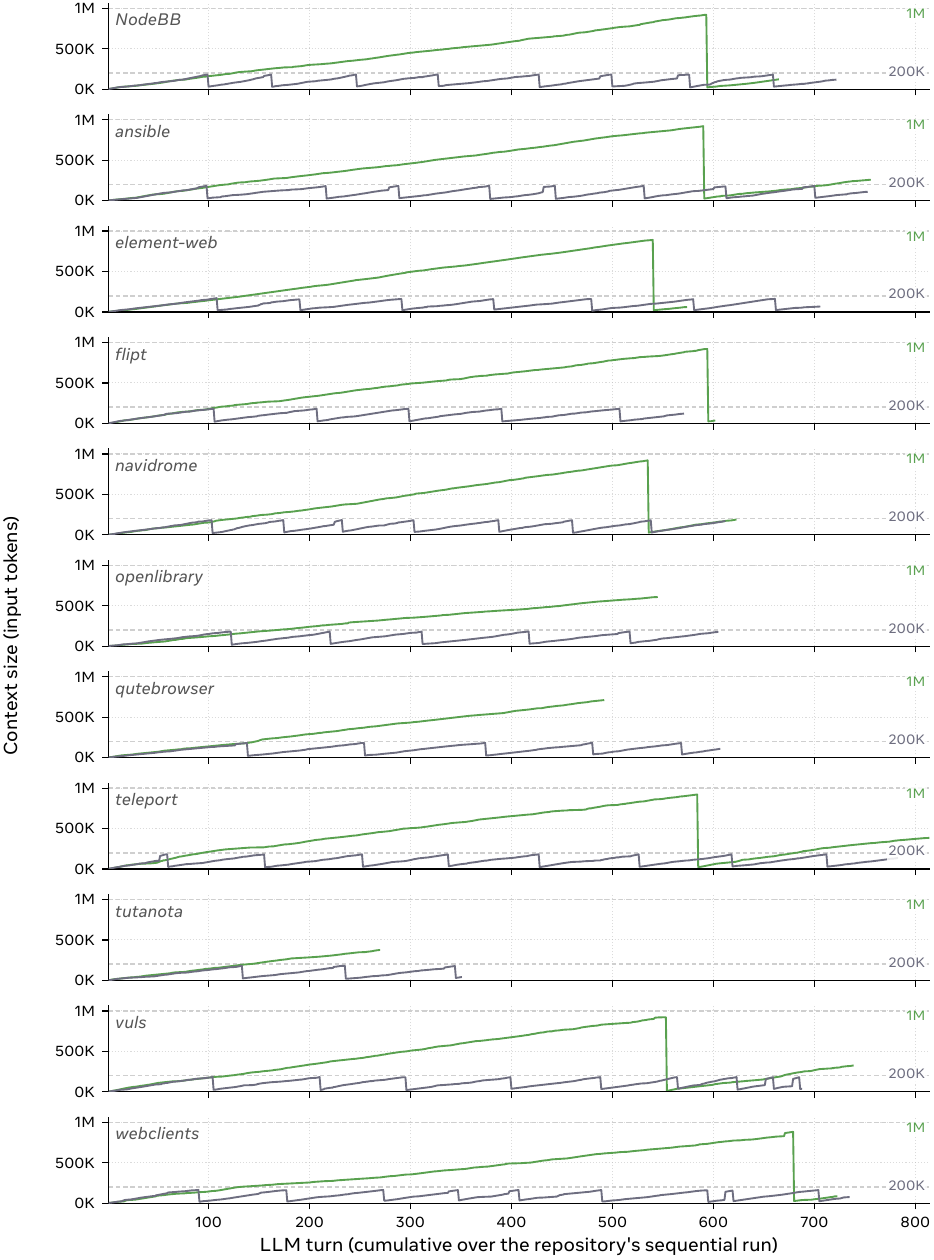}
  \caption{\textbf{\swe{} context evolution.} Context size (input tokens) per LLM turn under sequential evaluation for \gpt{}, one rollout per repository, with turns counted cumulatively over each repository's tasks. The 200K run is summarized frequently (median seven times per repository), whereas the 1M run (green) grows to about 900K tokens and is summarized at most once.}
  \label{fig:sbp-ctx-evolution}
\end{figure}

On \swe{}, sequences are long enough that the 200K context is summarized
repeatedly, whereas at 1M it retains nearly all history, growing to about
900K tokens (\autoref{fig:sbp-ctx-evolution}). Independent runs never
approach 200K, so they serve as the baseline for both windows. Keeping more
history does not help: at 1M, $\mathrm{avg@}3$, turns, and output tokens all fall
further for every model (\autoref{fig:swe-1m-main}), and the effort
reduction grows toward the end of each repository sequence
(\autoref{fig:swe-1m}). Together with the 32K results, these runs show that persistence effects
depend on how agents use accumulated context rather than how much of it fits
in the window.

Full-history summarizations within 200K \swe{} runs show how compaction and
adaptation interact directly. We compare each task against its own independent run, excluding the task that
triggers summarization. Effort reduction peaks in the two tasks immediately
before a summarization, when context holds 115K--160K tokens: \opus{}, \gpt{},
and \dps{} use 50\%, 78\%, and 29\% of their independent output tokens (52\%,
67\%, and 44\% of their turns). In the two tasks immediately after summarization, output effort
rebounds to 60\%, 94\%, and 46\% (67\%, 80\%, and 55\% of turns), while
pass-rate changes remain within noise (95\% CIs include zero). Compaction
therefore discards part of the history that adaptation relies on.

\FloatBarrier

\section{Seq-RL: details and additional results}
\label{app:seqrl-additional}
\suppressfloats[t]

This section gives the training data and settings
(\autoref{app:seqrl-training-details}), training-reward dynamics
(\autoref{app:seqrl-dynamics}), the downstream-credit ablation
(\autoref{app:seqrl-downstream-ablation}), and complete results
(\autoref{app:seqrl-complete-results}).

\subsection{Training details}
\label{app:seqrl-training-details}

\paragraph{Training data.}
The 27K training scenarios are human-authored and verified Search- and
Execution-style scenarios built following the ARE/\gaia{} annotation protocol
\citep{froger2025are} in the \gaia{} format: they use the same universe structure, apps, and tools, and
carry the same write-action DAG annotations and verifiers as the benchmark.
They come from universes disjoint from the evaluation set and are arranged into
1.6K compatible sequences of up to 50 scenarios with the procedure of
\autoref{app:benchmarks}. Each sequence yields one training example with six
trainable scenarios (\autoref{sec:seq-rl-setup}).

\paragraph{Local reward.}
For continuation $m$ at position $i$, let $y_{m,i}\in\{0,1\}$ denote task
success, and let $M$, $O$, and $E$ be the numbers of matched oracle
state-changing actions, oracle actions, and failure-causing extra actions. We
define
\begin{equation}
  p=\frac{M}{M+E},\qquad
  q=\frac{M}{O},\qquad
  F_2=\frac{5pq}{4p+q},\qquad
  r_{m,i}=(1-\alpha)y_{m,i}+\alpha F_2,
  \quad \alpha=0.15.
  \label{eq:seqrl-local-reward}
\end{equation}
We set $F_2=0$ when a denominator vanishes. When dense counts are unavailable,
the local reward falls back to $y_{m,i}$.

\paragraph{Downstream credit.}
The training reward mixes the local reward with the mean local reward of the
later scenarios in the same continuation:
\begin{equation}
  R_{m,i} =
  \begin{cases}
    \displaystyle
    \beta r_{m,i}
    +(1-\beta)\frac{1}{N-i}\sum_{j=i+1}^{N}r_{m,j},
      & i<N,\\[6pt]
    r_{m,N}, & i=N,
  \end{cases}
  \qquad \beta=0.85.
  \label{eq:seqrl-downstream-reward}
\end{equation}
Every trainable token produced for scenario $i$, including compaction tokens,
receives $R_{m,i}$. Missing judgments receive zero reward. A rollout is
discarded if more than half of its positions are unjudged, and truncated
rollouts are not graded.

\paragraph{Baseline and advantage.}
Let $L_{m,i}$ be the number of trainable tokens for scenario $i$ in
continuation $m$. Seq-RL uses one baseline for all continuations and positions
sharing a prefill:
\begin{equation}
  b=
  \frac{\sum_{m=1}^{G}\sum_{i=1}^{N}
    L_{m,i}~R_{m,i}}
  {\sum_{m=1}^{G}\sum_{i=1}^{N}L_{m,i}},
  \qquad
  A_{m,i}=R_{m,i}-b.
  \label{eq:seqrl-advantage}
\end{equation}
This pooled, token-weighted baseline replaces a position-wise one. Token
weighting, standard practice to keep response lengths from exploding, is the
only normalization: advantages are not divided by a standard deviation, and we
use no length penalty.

\paragraph{Configuration and compute.}
Each Seq-RL update contains $B=32$ prefill groups with $G=8$ continuations, or
256 rollouts of 6-scenario sequences. Training rollouts run in a lightweight internal harness with full-history
compaction and update the full model on 32 NVIDIA H200
nodes with a 128K-token context window, a learning rate of
$2.6\times10^{-6}$, temperature 1.0, and top-$p=1.0$. Rollouts are generated
asynchronously with a maximum policy lag of 16 trainer updates. Single-task RL
uses the same setup with no prefill and $N=1$: each of the 9.6K trainable
scenarios is sampled $G=8$ times per epoch from a fresh session, for three
epochs. All runs train on the exact same 9.6K scenarios (\autoref{sec:seq-rl-setup}). Because Seq-RL
rollouts are longer, each Seq-RL run ($\beta=0.85$ and $\beta=1$) takes about
25K H200 GPU-hours, Seq-RL without prefill about 22K, single-task RL with prefill
about 19K, and single-task RL about 15K.

\subsection{Training dynamics}
\label{app:seqrl-dynamics}

\begin{figure}[!ht]
  \centering
  \includegraphics[width=0.92\textwidth]{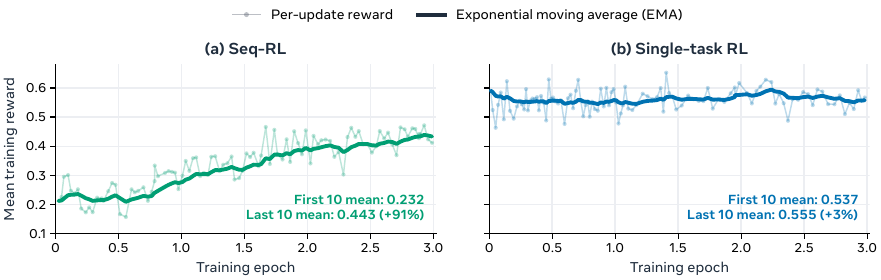}
  \caption{\textbf{Per-update local training reward under Seq-RL and single-task RL.}
  Light curves show the per-update mean local reward per scenario ($r_{m,i}$, \autoref{eq:seqrl-local-reward}) and solid curves their exponential moving averages. Annotated values report means over the first and last ten updates. Because Seq-RL scenarios follow a 100K-token prefill while single-task RL starts from fresh sessions, panels are shown separately to compare within-run progress rather than absolute height.}
  \label{fig:seqrl-reward-curves}
\end{figure}

\autoref{fig:seqrl-reward-curves} shows the per-update mean local reward behind
\autoref{sec:seqrl-training-dynamics}. Single-task RL starts higher because its
scenarios begin in a fresh session and stays near that level throughout
training (+3\%), whereas Seq-RL starts lower after a 100K-token prefill and
climbs steadily over the three epochs (+91\%). Because the two runs evaluate
local reward under different context distributions, we compare relative gain
within each run rather than absolute reward height.

\subsection{Downstream-credit ablation}
\label{app:seqrl-downstream-ablation}

\begin{figure}[!ht]
  \centering
  \includegraphics[width=0.92\textwidth]{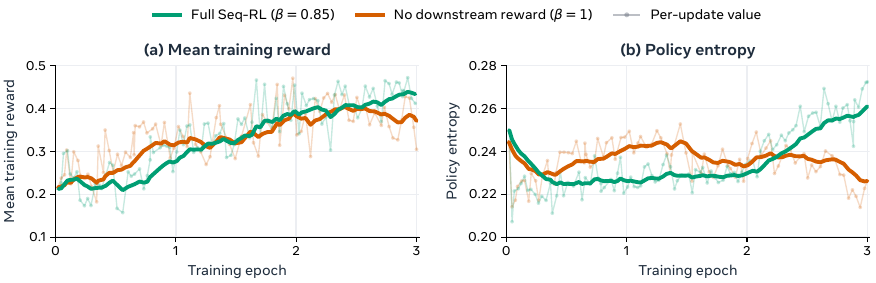}
  \caption{\textbf{Training dynamics with and without downstream credit.}
  Mean local reward per scenario (a), directly comparable between the two runs,
  and policy entropy (b) for full Seq-RL ($\beta=0.85$) versus local-reward-only
  ($\beta=1$). Light curves show per-update values and solid curves show
  exponential moving averages.}
  \label{fig:seqrl-ablation-dynamics}
\end{figure}

To isolate downstream credit, we compare full Seq-RL ($\beta=0.85$) with an
otherwise identical local-reward-only variant ($\beta=1$). With $\beta=1$,
the model trains on persistent histories but is rewarded only for the current
scenario, so nothing directly rewards leaving the context or environment
useful for later tasks. Its reward rises at first but plateaus earlier, and
its policy entropy declines after about two epochs, whereas with
$\beta=0.85$ the reward keeps rising and entropy rises above that of
$\beta=1$ late in training (\autoref{fig:seqrl-ablation-dynamics}). On held-out sequential \gaia{} and \swe{}, full Seq-RL scores higher in all
three evaluations (\autoref{tab:seqrl-downstream-ablation}). Because the 95\%
confidence intervals overlap, we treat this as limited evidence and keep
downstream credit in the final method.

\begin{table}[!ht]
  \centering
  \caption{\textbf{Effect of downstream credit on sequential \gaia{} and \swe{} performance.} Sequential evaluation at 200K ($K=5$ for \gaia{}, $K=3$ for \swe{}) after three epochs, in percent. $\pm$ gives the 95\% cluster-bootstrap half-width in gray (\autoref{app:eval-stats}), and $\beta=1$ removes downstream credit.}
  \label{tab:seqrl-downstream-ablation}
  \scriptsize
  \setlength{\tabcolsep}{8pt}
  \renewcommand{\arraystretch}{1.15}
  \newcommand{\ablationci}[2]{\ensuremath{#1{\color{black!50}\scriptstyle\,\pm#2}}}
  \begin{tabular}{l>{\hspace{16pt}}ccc}
    \toprule
      & \textbf{\gaia{} Search}
      & \textbf{\gaia{} Execution}
      & \textbf{\swe{}} \\
    \textbf{Objective}
      & \textbf{avg@5} & \textbf{avg@5} & \textbf{avg@3} \\
    \midrule
    \rowcolor{seqgreen!10}
    \textbf{Full Seq-RL ($\beta=0.85$)}
      & \ablationci{\boldsymbol{92.5}}{3.0}
      & \ablationci{\boldsymbol{86.3}}{6.1}
      & \ablationci{\boldsymbol{42.0}}{3.4} \\
    No downstream reward ($\beta=1$)
      & \ablationci{90.1}{3.0}
      & \ablationci{84.6}{4.2}
      & \ablationci{39.1}{3.3} \\
    \bottomrule
  \end{tabular}
\end{table}

\FloatBarrier
\subsection{Complete results}
\label{app:seqrl-complete-results}

\autoref{tab:seq-rl-main} gives full $\mathrm{avg@}K$, $\mathrm{pass@}K$,
and $\mathrm{pass}^{K}$ metrics at 200K for \muse-30B, single-task RL, and
Seq-RL. In the behavioral breakdowns, single-task RL emits the fewest output
tokens and turns sequentially on \gaia{} (\autoref{tab:seqrl-gaia2-behavior})
while suffering the largest sequential gaps. On \swe{}, \muse-30B and
single-task RL produce smaller patches sequentially as empty patches rise
(\autoref{fig:seqrl-behavior-position}(c)), whereas Seq-RL stays closer to its
independent patch size relative to the gold patch (\autoref{tab:seq-rl-swe-behavior}).

\begin{table}[!ht]
\centering
\caption{\textbf{Full metrics report for \muse{} checkpoints at 200K.} Values are percentages with $\Delta=\mathrm{Seq.}-\mathrm{Indep.}$ For $\mathrm{avg@}K$, $\pm$ gives the 95\% cluster-bootstrap half-width in gray (\autoref{app:eval-stats}), and other metrics are point estimates. \gaia{} uses $K=5$ and \swe{} $K=3$.}
\label{tab:seq-rl-main}
\scriptsize
\setlength{\tabcolsep}{2.2pt}
\renewcommand{\arraystretch}{1.03}

\newcommand{\bestscore}[1]{\boldsymbol{#1}}
\newcommand{\scoreci}[2]{#1{\color{black!50}\scriptstyle\,\pm#2}}

\begin{tabularx}{\linewidth}{@{}ll*{9}{>{\raggedleft\arraybackslash}X}@{}}
\toprule
& & \multicolumn{3}{c}{\textbf{\gaia{} Search}}
& \multicolumn{3}{c}{\textbf{\gaia{} Execution}}
& \multicolumn{3}{c}{\textbf{\swe{}}} \\
\cmidrule(lr){3-5}
\cmidrule(lr){6-8}
\cmidrule(lr){9-11}
\textbf{Metric} & \textbf{Method}
& \textbf{Indep.} & \textbf{Seq.} & $\Delta$
& \textbf{Indep.} & \textbf{Seq.} & $\Delta$
& \textbf{Indep.} & \textbf{Seq.} & $\Delta$ \\
\midrule

\multirow{3}{*}{avg@$K$}
& \muse-30B
& $\scoreci{91.4}{3.3}$
& $\scoreci{87.9}{4.2}$
& $\scoreci{-3.5}{4.0}$
& $\scoreci{86.1}{4.8}$
& $\scoreci{84.4}{4.3}$
& $\scoreci{-1.7}{3.8}$
& $\scoreci{45.8}{4.4}$
& $\scoreci{25.6}{4.0}$
& $\scoreci{-20.2}{2.8}$ \\

& + single-task RL
& $\scoreci{92.1}{3.3}$
& $\scoreci{87.9}{3.3}$
& $\scoreci{-4.3}{2.8}$
& $\scoreci{\bestscore{88.6}}{4.6}$
& $\scoreci{82.1}{4.6}$
& $\scoreci{-6.5}{5.9}$
& $\scoreci{46.3}{4.4}$
& $\scoreci{24.4}{2.9}$
& $\scoreci{-21.9}{2.9}$ \\

\rowcolor{seqgreen!10}
& \textbf{+ Seq-RL}
& $\scoreci{\bestscore{92.8}}{3.1}$
& $\scoreci{\bestscore{92.5}}{3.0}$
& $\scoreci{\bestscore{-0.3}}{2.3}$
& $\scoreci{87.2}{5.2}$
& $\scoreci{\bestscore{86.3}}{6.1}$
& $\scoreci{\bestscore{-0.9}}{7.0}$
& $\scoreci{\bestscore{48.0}}{4.5}$
& $\scoreci{\bestscore{42.0}}{3.4}$
& $\scoreci{\bestscore{-6.0}}{3.4}$ \\

\midrule
\multirow{3}{*}{pass@$K$}
& \muse-30B
& $\bestscore{97.5}$ & $95.6$ & $-1.9$
& $\bestscore{97.9}$ & $97.9$ & $0.0$
& $53.8$ & $37.6$ & $-16.2$ \\

& + single-task RL
& $\bestscore{97.5}$ & $96.3$ & $-1.3$
& $\bestscore{97.9}$ & $\bestscore{98.9}$ & $\bestscore{+1.0}$
& $55.5$ & $38.3$ & $-17.2$ \\

\rowcolor{seqgreen!10}
& \textbf{+ Seq-RL}
& $96.9$ & $\bestscore{97.5}$ & $\bestscore{+0.6}$
& $95.8$ & $95.8$ & $0.0$
& $\bestscore{56.2}$ & $\bestscore{53.1}$ & $\bestscore{-3.1}$ \\

\midrule
\multirow{3}{*}{pass$^K$}
& \muse-30B
& $79.4$ & $66.9$ & $-12.5$
& $66.3$ & $60.0$ & $-6.3$
& $36.0$ & $14.0$ & $-21.9$ \\

& + single-task RL
& $\bestscore{82.5}$ & $70.6$ & $-11.9$
& $\bestscore{74.7}$ & $57.9$ & $-16.8$
& $37.1$ & $11.4$ & $-25.7$ \\

\rowcolor{seqgreen!10}
& \textbf{+ Seq-RL}
& $81.9$ & $\bestscore{78.8}$ & $\bestscore{-3.1}$
& $70.5$ & $\bestscore{69.5}$ & $\bestscore{-1.0}$
& $\bestscore{41.0}$ & $\bestscore{28.8}$ & $\bestscore{-12.1}$ \\
\bottomrule
\end{tabularx}
\end{table}

\begin{table}[!ht]
  \centering
  \caption{\textbf{\gaia{} behavior of \muse{} checkpoints.} Per-scenario output tokens, LLM turns, and Python use (the percentage of scenario runs that call Python at least once), for Search and Execution.}
  \label{tab:seqrl-gaia2-behavior}
  \scriptsize
  \setlength{\tabcolsep}{2.4pt}
  \renewcommand{\arraystretch}{1.08}
  \begin{tabularx}{\linewidth}{@{}ll*{6}{>{\raggedleft\arraybackslash}X}@{}}
    \toprule
      & & \multicolumn{3}{c}{\textbf{Search}}
      & \multicolumn{3}{c}{\textbf{Execution}} \\
    \cmidrule(lr){3-5}\cmidrule(lr){6-8}
    \textbf{Evaluation} & \textbf{Checkpoint}
      & \shortstack{\textbf{Output}\\\textbf{tokens}}
      & \textbf{Turns}
      & \shortstack{\textbf{Python}\\\textbf{use}}
      & \shortstack{\textbf{Output}\\\textbf{tokens}}
      & \textbf{Turns}
      & \shortstack{\textbf{Python}\\\textbf{use}} \\
    \midrule
    \multirow{3}{*}{Independent}
      & \muse-30B   & $8.46\mathrm{K}$ & $39.1$ & $33.6\%$ & $5.99\mathrm{K}$ & $35.2$ & $12.1\%$ \\
      & + single-task RL & $5.25\mathrm{K}$ & $26.6$ & $43.9\%$ & $4.80\mathrm{K}$ & $36.2$ & $17.3\%$ \\
    \rowcolor{seqgreen!10}
      & \textbf{+ Seq-RL} & $6.70\mathrm{K}$ & $33.7$ & $80.3\%$ & $5.41\mathrm{K}$ & $34.6$ & $37.8\%$ \\
    \midrule
    \multirow{3}{*}{Sequential}
      & \muse-30B   & $3.96\mathrm{K}$ & $16.3$ & $67.0\%$ & $4.39\mathrm{K}$ & $23.8$ & $23.4\%$ \\
      & + single-task RL & $2.56\mathrm{K}$ & $10.4$ & $83.9\%$ & $3.38\mathrm{K}$ & $20.8$ & $47.2\%$ \\
    \rowcolor{seqgreen!10}
      & \textbf{+ Seq-RL} & $4.11\mathrm{K}$ & $16.5$ & $96.1\%$ & $4.20\mathrm{K}$ & $21.4$ & $76.4\%$ \\
    \bottomrule
  \end{tabularx}
\end{table}

\begin{table}[!ht]
  \centering
  \caption{\textbf{\swe{} behavior of \muse{} checkpoints.} Per-task statistics under independent and sequential evaluation at 200K. Patch/gold is the number of edited lines (additions plus deletions) in the agent patch relative to the gold patch, in percent.}
  \label{tab:seq-rl-swe-behavior}
  \scriptsize
  \setlength{\tabcolsep}{1.7pt}
  \renewcommand{\arraystretch}{1.08}
  \begin{tabularx}{\linewidth}{@{}ll*{6}{>{\raggedleft\arraybackslash}X}@{}}
    \toprule
      & & \multicolumn{3}{c}{\textbf{Effort}}
      & \multicolumn{3}{c}{\textbf{Patch statistics}} \\
    \cmidrule(lr){3-5}\cmidrule(lr){6-8}
    \textbf{Evaluation} & \textbf{Checkpoint}
      & \shortstack{\textbf{Output}\\\textbf{tokens}}
      & \textbf{\# Turns}
      & \shortstack{\textbf{\# Test}\\\textbf{turns}}
      & \shortstack{\textbf{Added}\\\textbf{lines}}
      & \textbf{Hunks}
      & \shortstack{\textbf{Patch/gold}\\\textbf{(\%)}} \\
    \midrule
    \multirow{3}{*}{Independent}
      & \muse-30B
      & $19.6\mathrm{K}$ & $94.0$ & $3.8$ & $97.8$ & $8.5$ & $81.8$ \\
      & + single-task RL
      & $14.8\mathrm{K}$ & $70.6$ & $3.5$ & $125.1$ & $9.0$ & $83.7$ \\
    \rowcolor{seqgreen!10}
      & \textbf{+ Seq-RL}
      & $18.8\mathrm{K}$ & $89.0$ & $4.4$ & $134.5$ & $9.1$ & $91.3$ \\
    \midrule
    \multirow{3}{*}{Sequential}
      & \muse-30B
      & $3.8\mathrm{K}$ & $22.0$ & $1.1$ & $73.6$ & $5.4$ & $61.0$ \\
      & + single-task RL
      & $3.6\mathrm{K}$ & $20.0$ & $1.2$ & $59.6$ & $5.7$ & $54.6$ \\
    \rowcolor{seqgreen!10}
      & \textbf{+ Seq-RL}
      & $8.7\mathrm{K}$ & $44.9$ & $2.7$ & $100.4$ & $8.2$ & $79.3$ \\
    \bottomrule
  \end{tabularx}
\end{table}

\autoref{tab:seq-rl-swe-repos} breaks down sequential \swe{} $\mathrm{avg@}3$ by
repository. Seq-RL scores highest of the three checkpoints in all 11
repositories, with gains of 2.5 to 27.5~pp over \muse-30B. Without the three
repositories with the largest gains, the task-weighted gain on the remaining
tasks is still 13.1~pp.

\begin{table}[!ht]
  \centering
  \caption{\textbf{Sequential \swe{} $\mathrm{avg@}3$ per repository for \muse{} checkpoints at 200K.} Values are percentages, with gains equal to Seq-RL minus each other checkpoint (pp). Each repository has 40 tasks, except tutanota (20).}
  \label{tab:seq-rl-swe-repos}
  \scriptsize
  \setlength{\tabcolsep}{7pt}
  \renewcommand{\arraystretch}{1.12}
  \begin{tabular}{@{}l@{\hspace{14pt}}rr>{\columncolor{seqgreen!10}}r@{\hspace{14pt}}rr@{}}
    \toprule
      & & & \multicolumn{1}{c}{} & \multicolumn{2}{c}{\textbf{Seq-RL gain vs.}} \\
    \cmidrule(l{10pt}){5-6}
    \textbf{Repository} & \textbf{\muse-30B} & \textbf{+ single-task RL} & \multicolumn{1}{r}{\textbf{+ Seq-RL}}
      & \textbf{\muse-30B} & \textbf{single-task RL} \\
    \midrule
    ansible     & 28.3 & 27.5 & \textbf{55.8} & $+27.5$ & $+28.3$ \\
    vuls        & 21.7 & 33.3 & \textbf{45.8} & $+24.2$ & $+12.5$ \\
    teleport    & 15.0 & 15.0 & \textbf{36.7} & $+21.7$ & $+21.7$ \\
    qutebrowser & 29.2 & 25.8 & \textbf{50.0} & $+20.8$ & $+24.2$ \\
    navidrome   & 13.3 & 15.0 & \textbf{33.3} & $+20.0$ & $+18.3$ \\
    flipt       & 20.0 & 27.5 & \textbf{38.3} & $+18.3$ & $+10.8$ \\
    element-web & 21.7 & 19.2 & \textbf{35.8} & $+14.2$ & $+16.7$ \\
    NodeBB      & 29.2 & 23.3 & \textbf{40.8} & $+11.7$ & $+17.5$ \\
    tutanota    & 21.7 & 20.0 & \textbf{31.7} & $+10.0$ & $+11.7$ \\
    openlibrary & 50.8 & 37.5 & \textbf{56.7} & $+5.8$  & $+19.2$ \\
    webclients  & 29.2 & 21.7 & \textbf{31.7} & $+2.5$  & $+10.0$ \\
    \bottomrule
  \end{tabular}
\end{table}

\end{document}